\documentclass[twocolumn,10pt]{asme2ej}
\usepackage{hyperref}   
\hypersetup{
	colorlinks=true,
	linkcolor=blue,
	citecolor=blue,
	urlcolor=blue,
}
\usepackage{epsfig}
\usepackage{amsmath}
\usepackage{amssymb}
\usepackage{graphicx}
\usepackage{subfigure}
\usepackage{verbatim}
\usepackage[thinlines,thiklines]{easybmat}
\usepackage{latexsym}
\usepackage[dvipsnames]{xcolor}
\usepackage{cite}
\usepackage{stmaryrd}
\usepackage[numbers,sort&compress]{natbib}
\usepackage{multirow}
\usepackage{textcomp}
\usepackage[ruled]{algorithm2e}
\usepackage{bm}
\usepackage{color,soul}
\sethlcolor{white}
\definecolor{shadecolor}{rgb}{1,0.9,0.4}

\renewcommand{\topmargin}{1.0in} 

\def\diag{\mathop{\rm diag}\nolimits}

\def\s{\mathop{\rm s}\nolimits}
\def\c{\mathop{\rm c}\nolimits}

\def\diag{\mathop{\rm diag}\nolimits}
\def\rank{\mathop{\rm rank}\nolimits}
\def\atan2{\mathop{\rm atan2}\nolimits}
\newcommand{\bs}[1]{\ensuremath{{\boldsymbol{#1}}}}
\newtheorem{thm}{Theorem}

\newtheorem{lem}{Lemma}

\def\atan2{\mathrm{atan2}}

\newcommand\highlightReference[1]{%
  \expandafter\newcommand\csname highlightReference-#1\endcsname{}%
}
\let\oldbibitem\bibitem
\def\bibitem#1 #2\par{%
  \expandafter\ifx\csname highlightReference-#1\endcsname\relax
    \oldbibitem{#1}#2\par
  \else
    \oldbibitem{#1}\highlight{#2}\par
  \fi
}
\newcommand\highlight[1]{\textcolor{blue}{#1}}

\newcommand\norm[1]{\left\lVert#1\right\rVert}

\newcounter{mynum1}

\def\s{\mathop{\rm s}\nolimits}
\def\c{\mathop{\rm c}\nolimits}

\def\diag{\mathop{\rm diag}\nolimits}
\def\rank{\mathop{\rm rank}\nolimits}
\def\spn{\mathop{\rm span}\nolimits}

\def\atan2{\mathrm{atan2}}

\begin{document}

\title{Gaussian Process-Based Learning Control of Underactuated Balance Robots with an External and Internal Convertible Modeling Structure}

\author{ Feng Han, Jingang Yi\thanks{Address all correspondence to J. Yi.}\affiliation{Department of Mechanical and Aerospace Engineering\\
Rutgers, The State University of New Jersey\\ Piscataway, NJ 08854, USA\\
Email: \{fh233, jgyi\}@rutgers.edu}}

\maketitle

\begin{abstract}
{\it \hspace{3mm}  External and internal convertible (EIC) form-based motion control is one of the effective designs of simultaneously trajectory tracking and balance for underactuated balance robots. Under certain conditions, the EIC-based control design however leads to uncontrolled robot motion. We present a Gaussian process (GP)-based data-driven learning control for underactuated balance robots with the EIC modeling structure. Two GP-based learning controllers are presented by using the EIC structure property. The partial EIC (PEIC)-based control design partitions the robotic dynamics into a fully actuated subsystem and one reduced-order underactuated system. The null-space EIC (NEIC)-based control compensates for the uncontrolled motion in a subspace, while the other closed-loop dynamics are not affected. Under the PEIC- and NEIC-based, the tracking and balance tasks are guaranteed and convergence rate and bounded errors are achieved without causing any uncontrolled motion by the original EIC-based control. We validate the results and demonstrate the GP-based learning control design performance using two inverted pendulum platforms.}
\end{abstract}

\section{Introduction}

An underactuated balance robot possesses fewer control inputs than the number of degrees of freedom (DOFs)~\cite{KANT2020Orbital, Han2023ICRA}. Motion control of underactuated balance robots requires both the trajectory tracking of the actuated subsystem and balance control of the unactuated, unstable subsystem~\cite{HanTMECH2022, ChenTRO2022, HanRAL2021}. Inverting the non-minimum phase unactuated nonlinear dynamics brings additional challenges in causal feedback control design. Several modeling and control methods have been proposed for these robots and their applications~\cite{Turrisi2022RAL, HanRAL2021, ChenTRO2022, BECKERS2019390,ChenIROS2015,GrizzleAUTO2014, Han2022Tmech}.  Orbital stabilization method was used for balancing underactuated robots~\cite{Shiriaev2005,CANUDASDEWIT2002527,Maggiore2013Virtual} with applications to bipedal robot~\cite{Chevall2009} and cart-invented pendulum~\cite{KANT2020Orbital}. A virtual constraint encoded the motion coordination between the actuated and unactuated subsystems. Energy shaping-based control was also designed for underactuated balance robots in~\cite{Fantoni2000,Xin2005TAC}. One feature of those methods is that the achieved balance-enforced trajectory is not unique and cannot be prescribed explicitly~\cite{Shiriaev2005,KANT2020Orbital}. In~\cite{GetzPhD,HanRAL2021}, a simultaneous trajectory tracking and balance control of underactuated balance robots was proposed by using the property of the external and internal convertible (EIC) form of the robot dynamics. The EIC-based control has been demonstrated as one of the effective approaches to achieve fast convergence and guaranteed performance that avoids non-causal continuous feedback control.

All of the above-mentioned control designs require an accurate model of robot dynamics and the control performance would deteriorate under model uncertainties or external disturbances. Machine learning-based methods provide an efficient tool for robot modeling and control. In particular, Gaussian process (GP) regression is an effective learning approach that generates nearly-analytical structure and bounded prediction errors~\cite{Lederer2023TAC, BECKERS2019390, Beckers2022TAC, pmlr2015Bach}. Development of GP-based performance-guaranteed control for underactuated balance robots has been reported~\cite{ChenTRO2022, Helwa2019RAL, Lederer2023TAC}. In~\cite{ChenTRO2022}, the control design was conducted into two steps. A GP-based inverse dynamics controller for unactuated subsystem to achieve balance and a model predictive control (MPC) was used to simultaneously track the given reference trajectory and estimate the balance equilibrium manifold (BEM). The GP prediction uncertainties were incorporated into the control design to enhance the control robustness. The work in~\cite{HanRAL2021} followed the sequential control design in the EIC-based framework and the controller was adaptive to the prediction uncertainties. The training data was selected to reduce the computational complexity.

This paper takes advantage of the structured GP modeling approach in~\cite{HanRAL2021, BECKERS2019390} and presents an integration of EIC-based control with GP models.  It is first shown that under certain conditions, there exist uncontrolled motions with the EIC-based control and this property can cause the entire system unstable. We identify these conditions and design the stable GP-based learning control with the properly selected nominal robot dynamic model. Two different controllers, called partial- and null-space-EIC (i.e., PEIC- and NEIC), are presented to improve the performance of the EIC-based control. The PEIC-based control constructs a virtual inertial matrix to re-shape the dynamics coupling interactions between the actuated and unactuated subsystems. The EIC-induced uncontrolled motion is eliminated and the robotic system behaves as a combined fully actuated subsystem and a reduced-order unactuated subsystem. Alternatively, the compensation effect in NEIC-based control is applied to the uncontrolled coordinates in the null space, while the other part of the stable system motion stays unchanged. The PEIC- and NEIC-based controls achieve guaranteed robust performance with a fast convergence rate.

The main contribution of this work lies in the new GP-based learning control of underactuated balance robots using the EIC structural properties. Compared with the approaches in~\cite{HanRAL2021, GetzPhD}, this work reveals underlying design properties and limitations of the EIC-based control for underactuated balance robots. We incorporate these discoveries into the GP-based data-driven modeling and learning control design. Such a GP-based control design has not been reported in previous work~\cite{HanRAL2021,ChenCASE2017, ChenICRA2019, ChenTRO2022, BECKERS2019390}. Compared with the work in~\cite{ChenICRA2019,ChenTRO2022}, the proposed method takes advantage of the attractive EIC modeling properties for control design and does not use MPC that requires highly computational demands. This paper is an extension of the previous conference submission~\cite{Han2024ICRA} with new design, analyses and experiments. Particularly, the NEIC-based control design and experiments are new in this paper.

The rest of the paper is outlined as follows. We discuss the EIC-based control for underactuated balance robots and present the problem statement in Section~\ref{Sec_Model}. Section~\ref{Sec_Nominal_Model} presents the GP-based data-driven robot dynamics. The PEIC- and NEIC-based controls are presented in Section~\ref{Sec_Control}. The stability analysis is discussed in Section~\ref{Sec_Stability}. The experimental results are presented in Section~\ref{Sec_Result} and finally, Section~\ref{Sec_Conclusion} summarizes the concluding remarks.

\begin{figure*}[th!]
	\hspace{-3mm}
	\subfigure[]{
	\label{Fig_EIC_diagram}
	\includegraphics[width=5.9cm]{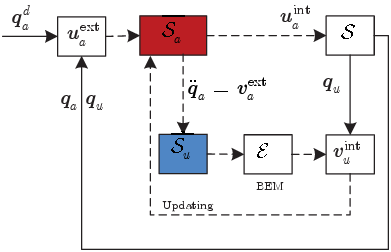}}
\hspace{-2mm}
	\subfigure[]{
		\label{Fig_PEIC}
		\includegraphics[width=5.8cm]{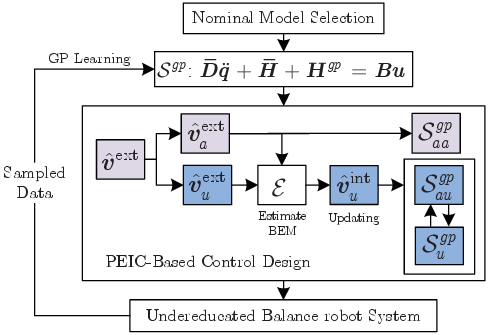}}
	\hspace{-2mm}
	\subfigure[]{
		\label{Fig_NEIC}
		\includegraphics[width=5.8cm]{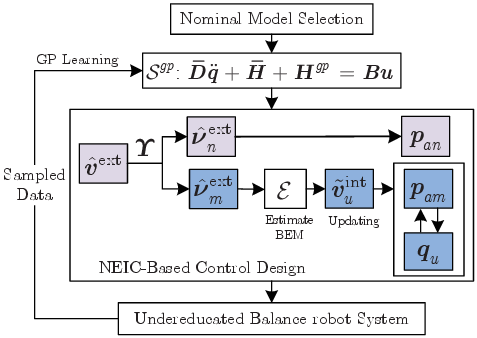}}
\vspace{-3mm}
\caption{Illustrative diagram of the control design for (a) the regular EIC-based control; (b) The PEIC-control design; and (c) the NEIC-based control design. In (a), the dashed line indicates the design flow and the solid line indicates the control flow.}
\label{Fig_Control_Diagram}
\vspace{-2mm}
\end{figure*}

\section{EIC-Based Robot Control and Problem Statement}
\label{Sec_Model}

\subsection{Robot Dynamics and EIC-Based Control }

We consider an underactuated balance robot with $(n+m)$ DOFs, $n, m\in \mathbb{N}$, and the generalized coordinates are denoted as $\bm q \in \mathbb{R}^{n+m}$. The robot dynamics is expressed as
\begin{equation}
	\label{Eq_Physical_Model}
     \mathcal S: \bm{D}(\bm{q})\ddot{\bm{q}} + \bm{C}(\bm{q},\dot{\bm{q}})\dot{\bm{q}} + \bm{G}(\bm{q}) = \bm{B}\bm{u},
\end{equation}
where $\bm D(\bm q)$, $\bm{C}(\bm{q},\dot{\bm{q}})$ and $\bm{G}(\bm q)$ are the inertial matrix, Coriolis, and gravity matrix, respectively. $\bm B$ denotes the input matrix and $\bm u \in \mathbb{R}^n$ is the control input. The coordinates are partitioned as $\bm q=[\bm q_a^T\; \bm q_u^T]^T$, with actuated coordinate $\bm q_a \in \mathbb{R}^{n}$ and unactuated  coordinate $\bm q_u \in \mathbb{R}^m$. We focus on the case $n \geq m$, and without loss of generality, we assume that $\bm B=[\bm I_n~\bm 0]^T$, where $\bs{I}_n \in \mathbb{R}^n$ is the identity matrix with dimension $n$. The robot dynamic model in~\eqref{Eq_Physical_Model} is rewritten as
\begin{subequations}
\label{Eq_Sub_Dynamics}
\begin{align}
    \mathcal S_a&: \bm D_{a a} \ddot{\bm q}_{a}+\bm D_{a u} \ddot{\bm q}_{u}+\bm H_{a}=\bm u,\label{Eq_Actuated}\\
     \mathcal S_u&:\bm D_{u a} \ddot{\bm q}_{a}+\bm D_{u u} \ddot{\bm q}_{u}+\bm H_{u}=\bm 0 \label{Eq_unactuated}
\end{align}
\end{subequations}
for actuated ($\mathcal S_a$) and unactuated ($\mathcal S_u$) subsystems, respectively. Subscripts ``$aa$ ($uu$)'' and ``$ua$ and $au$'' indicate the variables related to the actuated (unactuated) coordinates and coupling effects, respectively. For presentation convenience, we introduce $\bm H=\bm{C}\dot{\bm{q}} + \bm{G}$, $\bm H_a = \bm C_{a} \dot{\bm q}+\bm G_{a}$, and $\bm H_u=\bm C_{u} \dot{\bm q}+\bm G_{u}$, and the dependence of $\bs{D}$, $\bm C$, and $\bm G$ on $\bm q$ and $\dot{\bm q}$ is dropped. Subsystems $\mathcal S_a$ and $\mathcal S_u$ are also referred to as the external and internal subsystems, respectively~\cite{GetzPhD,ChenTRO2022}.

The control goal is to steer actuated coordinate $\bm q_a$ to follow a given desired trajectory $\bm q_a^d$ for $\mathcal S_{a}$, while the unactuated, unstable subsystem $\mathcal S_{u}$ is balanced at unknown equilibrium $\bm q_u^e$. Obtaining the estimate of $\bm q_u^e$ needs to invert the non-minimum phase dynamics $\mathcal S_{u}$, which is challenging for non-causal control design.

The EIC-based control in~\cite{GetzPhD,HanRAL2021} first designs external input $\bs{u}$ to follow $\bm q_a^d$ by temporarily neglecting $\mathcal S_{u}$, namely,
\begin{equation}
\label{Eq_Actuated_Control}
\bm u^\mathrm{ext}= \bm{D}_{a a} \bm v^\mathrm{ext}+\bm{D}_{a u} \ddot{\bm q}_{u}+\bm H_{a},
\end{equation}
where $\bm v^\mathrm{ext}=\bm q_a^d-\bs{k}_{d1} \dot{\bs{e}}_a-\bs{k}_{p1} {\bs{e}}_a$ is the auxiliary input under which tracking error $\bm e_a=\bs{q}_a-\bm q_a^d$ converges to zero, $\bs{k}_{p1},\bs{k}_{d1}$ are diagonal matrices with positive elements. To account for the coupling effect between $\mathcal S_{a}$ and $\mathcal S_{u}$, the BEM is introduced as the equilibrium of $\bm q_u$ under $\bm v^\mathrm{ext}$, namely,
\begin{equation}
\label{Eq_BEM}
\mathcal{E}=\left\{\bm q_u^{e}:  \bm \Gamma\left(\bm q_u;\bm v^\mathrm{ext}\right)=\bm 0, \dot{\bm q}_u=\ddot{\bm q}_u=\bm 0\right\},
\end{equation}
where $\bm \Gamma(\bm q_u;\bm v^\mathrm{ext})=\bm{D}_{uu} \ddot{\bm q}_u + \bm D_{ua} \bm v^\mathrm{ext}+ \bm H_{u}$. $\bm q_u^e$ is obtained by inverting $\bm \Gamma_0=\bm \Gamma(\bm q_u;\bm v^\mathrm{ext})\vert_{ \dot{\bm q}_u=\ddot{\bm q}_u=\bm 0}=\bm 0$.

To stabilize $\bm q_u$ onto $\mathcal{E}$, the control design updates $\bm q_a$ balance motion as
\begin{equation}
\label{Eq_Va_Int}
\bm v^\mathrm{int}=-\bm{D}_{u a}^{+}(\bm H_{u}+\bm D_{uu} \bm v_u^\mathrm{int}),
\end{equation}
where $\bm{D}_{u a}^{+}=(\bm{D}^T_{u a}\bm{D}_{u a})^{-1}\bm{D}^T_{u a}$ is the generalized inverse of $\bm{D}_{u a}$, $\bm v_u^\mathrm{int}=\bm q_u^e-\bs{k}_{d2} \dot{\bs{e}}_u-\bs{k}_{p2} {\bs{e}}_u$ is the auxiliary control that drives error $\bm e_u=\bs{q}_u-\bm q_u^e$ towards zero, and $\bs{k}_{p2},\bs{k}_{d2}$ are diagonal matrices with positive elements. The final control is obtained by replacing $\bm v^\mathrm{ext}$ in~\eqref{Eq_Actuated_Control} with $\bm v^\mathrm{int}$ in~\eqref{Eq_Va_Int}, that is,
\begin{equation}\label{Eq_Ua_Int}
  \bm u^\mathrm{int} =  \bm{D}_{aa} \bm v^\mathrm{int}+\bm{D}_{a u} \ddot{\bm q} _{u}+\bm{H}_{a}.
\end{equation}

Fig.~\ref{Fig_EIC_diagram} illustrates the above sequential EIC-based control design. It has been shown in~\cite{GetzPhD} that the control $\bm u^\mathrm{int}$ guarantees both $\bm e_a$ and $\bm e_u$ convergence to a neighborhood of origin exponentially and therefore, the EIC-based control achieves trajectory tracking for $\mathcal S_{a}$ and balancing task for $\mathcal S_{u}$ simultaneously.

\subsection{Motion Property under EIC-based Control}

Control design~\eqref{Eq_Va_Int} uses a mapping from low-dimensional ($m$) to high-dimensional ($n$) space (i.e., $n \geq m$). Under control~\eqref{Eq_Ua_Int}, it has been shown in~\cite{GetzPhD} that there exists a finite time $T>0$ and for small number $\epsilon>0 $, $\norm{\bm q_u(t)- \bm q_u^{e}(t)}<\epsilon$ for $t>T$. Therefore, given the negligible error, we obtain $\bm D_{ua}(\bm q_a, \bm q_u) \approx  \bm D_{ua}(\bm q_a, \bm q_u^{e})$.

For~$\mathcal S$ in~\eqref{Eq_Sub_Dynamics}, if $\rank(\bm D_{au})=m$ for all $\bm q$, applying singular value decomposition (SVD) to $\bm D_{ua}$ and $\bm D_{ua}^+$, we have
\begin{align}
\label{Eq_Dua_SVD}
\bm D_{ua}=\bm U\bm \Lambda \bm V^T, \quad \bm D_{ua}^{+}=\bm V \bm \Lambda^+ \bm U^T,
\end{align}
where $\bm U=[\bm u_1~\cdots~\bm u_m] \in \mathbb{R}^{m \times m}$ and $\bm V \in \mathbb{R}^{n\times n}$ are unitary orthogonal matrices. $\bm \Lambda =[\bm \Lambda_m ~\bm 0] \in\mathbb{R}^{m \times n}$, $ \bm \Lambda^+ =[\bm \Lambda_m^{-1}~\bm 0]^T\in\mathbb{R}^{n\times m}$ and $\bm \Lambda_m=\diag(\sigma_{1},\cdots,\sigma_{m})$ with singular values $\sigma_i>0$, $i=1,\cdots,m$. We partition $\bm V$ into the block matrix $\bm V=[\bm V_{m}~\bm V_{n}]$, $\bm V_{m} \in \mathbb{R}^{n\times m}$ and $\bm V_{n} \in \mathbb{R}^{n\times {(n-m)}}$. Since $\rank(\bm D_{au})=m$, the null space of $ \bm D_{ua}$ is $\ker(\bm D_{ua})=\spn\{\bm V_{n}\}$.

Column vectors of matrix $\bm V$ serve as a basis in $\mathbb{R}^{n}$ and we introduce a coordinate transformation $\bm \Upsilon: \bs{x} \mapsto \bm V^T\bm x$ for $\bm x \in \mathbb{R}^{n}$. Clearly, $\bm \Upsilon$ is a linear, time-varying, smooth map. Applying $\bm \Upsilon$ to $\bm q_a$ and $\bm v^\mathrm{ext}$, we have
\begin{equation}
\bm p_a=\bm V^T \bm q_a, \quad \bm \nu^\mathrm{ext}=\bm V^T \bm v^\mathrm{ext},
\label{Eq_transform}
\end{equation}
where $\bm p_a=[\bm{p}_{am}^T\;\bs{p}_{an}^T]^T$, $\bm \nu^\mathrm{ext}=[(\bm \nu_m^\mathrm{ext})^T\;(\bm \nu_n^\mathrm{ext})^T]^T$, and $\bs p_{am}, \bm \nu_m^\mathrm{ext} \in\mathbb{R}^{m}$, $\bs p_{an}, \bm \nu_n^\mathrm{ext}\in\mathbb{R}^{n-m}$. Note that $[\bm p_a^T~\bm q_u^T]^T$ still serves as a complete set of generalized coordinates for $\mathcal S$. Using the new coordinate $\bm p_a$, we have the following motion property under the EIC-based control for $\mathcal S$ and the proof is given in Appendix~\ref{proof_free_motion}.

\begin{lem}
\label{Lem_Uncontrolled_Motion}
For $\mathcal S$ in~\eqref{Eq_Sub_Dynamics}, if $\rank(\bm D_{au})=m$ holds for $\bm q$ and all $n$ control inputs appear in $\mathcal S_u$ dynamics (through $\ddot{\bs{q}}_a$), under the EIC-based control~\eqref{Eq_Ua_Int}, the BEM $\mathcal{E}$ in~\eqref{Eq_BEM} is associated with only $\bm \nu_{m}^\mathrm{ext}$ and robot dynamics can be written into
\begin{subequations}\label{Eq_Su_Vu2}
\begin{align}
\hspace{-3mm}
\mathcal S_\mathrm{EIC}:~
&\ddot p_{ai}=-\frac{\bm u_{i}^T\left(\bm H_u+\bm D_{uu}  \bm v_u^{\mathrm{int}}\right)}{\sigma_{i}},\;\,i=1,\cdots,m,\label{Eq_Su_Vu2-a} \\
&\ddot p_{aj}=0, \;\,j=m+1,\cdots,n,\label{Eq_Su_Vu2-b} \\
&\ddot {\bm q}_{u} = \bm v_{u}^\mathrm{int}.
\end{align}
\end{subequations}
\end{lem}

No control input appears for coordinates in $\ker(\bm D_{ua})$ as shown in~\eqref{Eq_Su_Vu2-b} and only $m$ actuated coordinates in $\spn(\bm V)$ are under active control, as shown in~\eqref{Eq_Su_Vu2-a}. The results in Lemma~\ref{Lem_Uncontrolled_Motion} reveal the motion property of the EIC-based control design. The uncontrolled motion happens to a special set of underactuated balance robots under the conditions in Lemma~\ref{Lem_Uncontrolled_Motion}. If the unactuated motion is only related to $m$ out of $n$ control inputs, the motion~\eqref{Eq_Su_Vu2-b} vanishes and the EIC-based control works well. In~\cite{HanRAL2021}, the EIC-based control worked properly for the rotary inverted pendulum with $n=m=1$. In~\cite{ChenTRO2022,WangJDSMC2023}, the EIC-based control also worked for the bikebot with $n=2$ (planar motion) and $m=1$ (roll motion) but the roll motion depends on steering control only, that is, no velocity control, and therefore, does not satisfy the condition for Lemma~\ref{Lem_Uncontrolled_Motion}. We will show an example of the 3-link inverted pendulum platform that demonstrates the uncontrolled motion under the EIC-based control in Section~\ref{Sec_Result}.

%


With the above-discussed motion property under the EIC-based control, we consider the following problem.

\emph{Problem Statement}: The goal of robot control is to design an enhanced EIC-based learning control to drive the actuated coordinate $\bs{q}_a$ to follow a given profile $\bs{q}_a^d$ and simultaneously the unactuated coordinate $\bs{q}_u$ to be stabilized on the estimated profile $\bs{q}_u^e$ using the GP-based data-driven model. The uncontrolled motion presented in Lemma~\ref{Lem_Uncontrolled_Motion} should be avoided for robot dynamics~\eqref{Eq_Sub_Dynamics}.

\vspace{-2mm}
\section{GP-Based Robot Dynamics Model}
\label{Sec_Nominal_Model}

We build a GP-based robot dynamics model that will be used for control design in the next section.

\vspace{-2mm}
\subsection{GP-Based Robot Dynamics Model}

To keep self-contained, we briefly review the GP model. We consider a multivariate continuously smooth function $y=  f(\bm x) + w$, $\bm x_i\in \mathbb{R}^{n_x}$, where $w \in \mathbb{R}$ is the zero-mean Gaussian noise and $n_x$ is the dimension of $\bs{x}$. Denote the training data as $\mathbb D=\{\bm{X}, \bm{Y}\}=\left\{\bm  x_i, y_{i}\right\}_{i=1}^N$, where $\bm{X}=\{\bm x_i\}_{i=1}^N$, $\bm{Y}=\{y_i \} _{i=1}^N$, and $N \in \mathbb{N}$ is the number of the data point. The GP model is trained by maximizing posterior probability $p(\bm Y; \bm X, \bm \Theta)$ over the hyperparameters $\bm \Theta$, that is, $\bm \Theta$ is obtained by solving
\begin{equation*}
  \min_{\bm \Theta} -\log(\bm Y;\bm X, \bm \Theta) =\min_{\bm \Theta} -\frac{1}{2} \bm{Y}^{T}\bm{K}^{-1} \bm{Y}  -\frac{1}{2}\log\det(\bm K),
\end{equation*}
where $\bm K=(K_{ij})$, $K_{ij}=k(\bm x_i, \bm x_j)=\sigma_{f}^{2} \exp (-\frac{1}{2}(\bm x_i-\bm x_j)^{T} \bm{W}(\bm x_i-\bm x_j))+\vartheta^2\delta_{ij}$, $\bm W = \diag\{W_1,\cdots, W_{n_x}\}>0$, $\delta_{ij}=1$ for $i=j$, and $\bm \Theta =\{\bm W, \sigma_f, \vartheta\}$ are hyperparameters.

Given $\bm  x^*$, the GP model predicts the corresponding $y$ and the joint distribution is
\begin{equation}
\begin{bmatrix} \bm Y \\  y \end{bmatrix} \sim\mathcal{N}
\left(\bm 0, \begin{bmatrix} \bm K & \bm k^T \\ \bm k & k^*   \end{bmatrix}
\right),
\end{equation}
where $\mathcal{N}(\bs{\mu},\bs{\Sigma})$ denotes the Guassian distribution with mean $\bs{\mu}$ and variance $\bs{\Sigma}$, $\bm{k}=\bm k(\bm x^*, \bm X)$ and $k^*= k(\bm x^*, \bm x^*)$. The mean value and variance for input $\bm x^*$ are
\begin{equation}\label{Eq_GP_Pred}
    \mu_i(\bm  x^*) = \bm{k}^{T}\bm{K}^{-1}\bm Y,~ \Sigma_i(\bm x^*)=k^{*} - \bm{k}\bm{K}^{-1} \bm{k}^{T}.
\end{equation}

For robot dynamics $\mathcal{S}$ in~\eqref{Eq_Physical_Model}, we first build a nominal model
\begin{equation}
\label{Eq_nominal_model}
\mathcal S^n: \; \bar{\bm D}\ddot{\bm q} +\bar{\bm H}=\bm B \bm u,
\end{equation}
where $\bar{\bm D}$ and $\bar{\bm H}$ are the nominal inertia and nonlinear matrices, respectively. In general, the nominal dynamic model does not hold for the data sampled from the physical robot systems. The GP models are built to capture the difference between $\mathcal S^n$ and $\mathcal S$, namely,
\begin{equation*}
\bm H^e =\bm D\ddot{\bm q}+\bm H -\bar{\bm D}\ddot{\bm q}-\bar{\bm H}=\bm B \bm u-\bar{\bm D}\ddot{\bm q}-\bar{\bm H}.
\end{equation*}
We build GP models to estimate  $\bm H^e=[(\bm H^e_{a})^T \; (\bm H^e_{u})^T]^T$, where $\bm H^e_a$ and $\bm H^e_u$ are for $\mathcal{S}_a$ and $\mathcal{S}_u$, respectively. The training data $\mathbb D=\{\bs{X},\bs{Y}\}$ are sampled from $\mathcal S$ as $\bm{X}=\{\bm q,\,\dot {\bm q},\,\ddot {\bm q}\}$ and $\bm{Y}=\{\bm H^e\}$.

The GP predicted mean and variance are denoted as $(\bm \mu_i(\bm x),\bs{\Sigma}_i(\bm x))$ for $\bm H^e_i$, $i=a,u$. The GP-based robot dynamics models $\mathcal{S}^{gp}_a$ and $\mathcal{S}^{gp}_u$ for $\mathcal{S}_a$ and $\mathcal{S}_u$ respectively are given as
\begin{subequations}
\label{Eq_gp_model}
\begin{align}
\mathcal S_a^{gp}: & \; \bar{\bm D}_{aa}\ddot{\bm q}_a +\bar{\bm D}_{au}\ddot{\bm q}_u +\bm H_a^{gp}=\bm u, \label{Eq_gp_model:a}\\
\mathcal S_u^{gp}: &\;\bar{\bm D}_{ua}\ddot{\bm q}_a+\bar{\bm D}_{uu}\ddot{\bm q}_u +\bm H_u^{gp}=\bm 0, \label{Eq_gp_model:b}
\end{align}
\end{subequations}
where $\bm H_i^{gp} =\bar{\bm H}_i+\bm \mu_i(\bm x)$, $i=a,u$. The GP-based model prediction error is
\begin{equation}
\bs{\Delta}=\begin{bmatrix} \bm \Delta_a \\ \bm \Delta_u\end{bmatrix}=\begin{bmatrix} \bm \mu_a(\bm x)-\bm H^e_a \\ \bm \mu_u(\bm x)-\bm H^e_u \end{bmatrix}.
\label{modelerror}
\end{equation}
To quantify the GP-based model prediction error, the following property for $\bm \Delta$ is obtained directly from Theorem~6 in~\cite{GPtheory}.
\begin{lem}
\label{Lemma_GP_Errpr}
Given training dataset $\mathbb D$, if the kernel function $k(\bm x_i, \bm x_j)$  is chosen such that $\bm H^e_a$ for $\mathcal S_a$ has a finite reproducing kernel Hilbert space norm $\norm{\bm H^e_a}_{k}<\infty$, for given $0<\eta_a<1$,
\begin{equation}
\label{Eq_GP_Error}
  \Pr\left\{\norm{\bs{\Delta}_a } \leq \|\bm \kappa_a ^T\bm \Sigma_a^{1/2}(\bm  x)\| \right\} \geq \eta_a,
\end{equation}
where $\Pr\{\cdot\}$ denotes the probability of an event, $\bm \kappa_a \in \mathbb R^{n}$ and its $i$-th entry is $\kappa_{ai}=\sqrt{2\|{H}^e_{a,i}\|_{k}^{2}+300 \varsigma_i \ln ^{3} \frac{N+1}{1-\eta_a^{\frac{1}{n}}}}$,  $\varsigma_i=\max_{\bm x, \bm  x^{\prime} \in \bm{X}} \frac{1}{2} \ln | 1 +\vartheta_i^{-2} k_i\left(\bm  x, \bm  x^{\prime}\right) |$. A similar conclusion holds for $\bs{\Delta}_u$ with $0<\eta_u<1$.
\end{lem}

\vspace{-1mm}
\subsection{Nominal Model Selection}

The nominal model plays an important role in the EIC control. We consider the following conditions for choosing the nominal GP model $\mathcal{S}^{gp}$ to overcome the uncontrolled motion that was pointed out in Lemma~\ref{Lem_Uncontrolled_Motion} under the learning control.
\begin{list}{$\mathcal{C}_{\arabic{mynum1}}$:}{\usecounter{mynum1} \parsep 0.05in \leftmargin 0.3in \labelwidth 0.2in}
  \item $\bar{\bm D}=\bar{\bm D}^T$ is positive definite, $\norm {\bar{\bm D}}\le d$, $\norm {\bar{\bm H}}\le h$, where constants $0< d, h < \infty $;
\item $\rank(\bar{\bm D}_{aa})=n$, $\rank(\bar{\bm D}_{uu})=\rank(\bar{\bm D}_{ua})=m$; and
\item non-constant kernel of $\bar{\bm D}_{ua}$.
\end{list}

With $\mathcal C_1$ and $\mathcal C_2$, the generalized inversions of $\bar{\bm D}_{aa}$, $\bar{\bm D}_{uu}$, and $\bar{\bm D}_{au}$ exist, which are used to compute the auxiliary controls. We can select $\bar{\bm D}=\bar{\bm D}^T$ to ensure $\bar{\bm D}_{au}=\bar{\bm D}_{ua}^T$. To see the requirement of $\mathcal C_3$, we rewrite $\bm q_a =\sum_{i=1}^{n} p_{ai}\bm v_{i} $. By~\eqref{Eq_Su_Vu2}, under the updated control $\bm v^\mathrm{int}$, $\ddot{\bm q}_a = \sum\nolimits_{i=1}^{m} \ddot p_{ai}\bm v_{i}+ \sum\nolimits_{i=m+1}^{n} \ddot p_{ai}\bm v_{i}$, where $\bm v_{i}$ is the $i$th column of $\bs{V}$. Note that the part $\sum_{i=m+1}^{n} \ddot p_{ai}\bm v_{i}$ of $\mathcal S_a$ dynamics is free of control if $\bs{V}$ is constant. In spite of the fact that $\bm q_u$ stabilizes on $\bm q_u^{e}$, $\bm q_a$ converges to $\bm q_a^d$ only in an $m$-dimensional subspace and the other $(n-m)$ dimensional motion uncontrolled. If the system is stable, the uncontrolled motion cannot be fixed in the configuration space throughout entire control process.  Therefore, a non-constant kernel $\bar{\bm D}_{ua}$ is needed.

Conditions~$\mathcal C_1$-$\mathcal C_3$ provide the sufficient control-oriented nominal model selection criteria. The commonly used nominal model in~\cite{BECKERS2019390,HanRAL2021} is $\bar{\bm D}\ddot{\bm q}=\bm B \bm u$ with $\bar{\bm H}=\bm 0$.  The constant nominal model is used in~\cite{BECKERS2019390} as the system is fully actuated. It is not difficult to satisfy the above nominal model conditions in practice. First, the nonlinear term is canceled by feedback linearization and $\bar{\bm H}=\bm 0$ can be used. Matrix $\bar{\bm D}$ captures the robots' inertia property. The mass and length of robot links are usually available or can be measured. Meanwhile, dynamics coupling for revolute joints shows up in the inertia matrix as trigonometric functions of the relative joint angles. Therefore, the diagonal elements can be filled with mass or inertia estimates and the off-diagonal entries can be constructed with trigonometric functions multiplying inertia constants. We will show model selection examples in Section~\ref{Sec_Result}.

\section{GP-Enhanced EIC-Based Control}
\label{Sec_Control}

In this section, we propose two enhanced controllers using the GP model $\mathcal S^{gp}$, i.e., PEIC- and NEIC-based control. The PEIC-based control aims to eliminate uncontrolled motion under the EIC-based control, while the NEIC-based control directly manages the uncontrolled motion.

\subsection{Robust Auxiliary Control}

The control design should follow the guideline: (1) the $\bm p_{am}$ and $\bm q_{u}$ dynamics are preserved (since they are stable under the EIC-based control) and (2) the uncontrolled motion (in $\mathcal S^{gp}_{a}$) is either eliminated or under active control. The second requirement also implies that the motion of the unactuated coordinates depends on only $m$ control inputs. To see this, solving $\ddot{\bm q}_a$ from $\mathcal S^{gp}_{a}$ and plugging it into  $\mathcal S^{gp}_{u}$ yields
\begin{align*}
    (\bar{\bm D}_{u u}-\bar{\bm D}_{u a} \bar{\bm D}_{a a}^{-1} \bar{\bm D}_{a u})\ddot{\bm q}_u+ \bm{H}_{u}^{gp}-\bar{\bm D}_{u a} \bar{\bm D}_{a a}^{-1} \bm{H}_{a}^{gp}=-\bar{\bm D}_{u a} \bar{\bm D}_{a a}^{-1} \bm u.
\end{align*}
Note that $\bar{\bm{D}}_{ua}\in \mathbb{R}^{m\times n}$, $\bar{\bm{D}}_{a a}^{-1}\in\mathbb{R}^{n\times n}$, and $\bm q_u$ is overactuated given $n=\dim(\bm u)\geq m=\dim(\bm q_u)$. If $\bm q_u$ depends on the same number of control inputs, $(n-m)$ column vectors in $\bar{\bm{D}}_{u a} \bar{\bm{D}}_{a a}^{-1}$ should be zero. Thus, the EIC-based control is applied between the same number of actuated and unactuated coordinates.

With $\mathcal S^{gp}$, we update the predictive variance of $\mathcal S^{gp}_a$ into the auxiliary control $\bm v^\mathrm{ext}$ in~\eqref{Eq_Actuated_Control} as
\begin{equation}
\hat{\bm v}^\mathrm{ext} = \ddot{\bm q}_a^d - \hat{\bm k}_{p1}\bm e_a- \hat{\bm k}_{d1}\dot{\bm e}_a
\label{control1}
\end{equation}
where $\hat{\bm k}_{p1}=\bs{k}_{p1}+k_{n1}\bm \Sigma_a$ and $\hat{\bm k}_{d1}=\bs{k}_{d1}+k_{n2}\bm \Sigma_a$ are control gains with parameters $k_{n1}, k_{n2} \geq 0$. The variance of GP prediction $\bm \Sigma_a$ captures the uncertainty in robot dynamics and is updated online with sensor measurements.

Given the GP-based dynamics, the BEM is estimated by solving the optimization problem
\begin{equation}
\label{Eq_BEM_opt}
\hat{\bm q}_u^e=\arg \min_{\bm q_u} \| \bs{\Gamma}_0 (\bm q_u;\hat{\bm v}^\mathrm{ext})\|.
\end{equation}
The updated control design is
\begin{equation}
    \hat{\bm v}_u^\mathrm{int} = \ddot {\hat{\bm q}}_u^{e} -\hat{\bm k}_{p2}\hat{\bm e}_u - \hat{\bm k}_{d2}\dot{\hat{\bm e}}_u,
\label{control2}
\end{equation}
where $\hat{\bm e}_u =\bm q_u- \hat{\bm q}_u^{e}$ is the internal system tracking error relative to the estimated BEM. Similar as $\hat{\bm k}_{p2}, \hat{\bm k}_{d2}$,  $\hat{\bm k}_{p2}=\bs{k}_{p2}+k_{n3}\bm \Sigma_u$ and $\hat{\bm k}_{d2}=\bs{k}_{d2}+k_{n4}\bm \Sigma_u$ depend on $\bm \Sigma_u$ with the parameters by $k_{n3 }, k_{n4} \geq 0$.

Let $\Delta \bm q_u^e=\bm q_u^e-\hat{\bm q}_u^e$ denote the BEM estimation error and the actual BEM is $\bm q_u^e =\hat{\bm q}_u^ {e}+\Delta \bm q_u^e$. The control design based on actual BEM should be $\bm v_u^\mathrm{int} = \ddot {\bm q}_u^e -\hat{\bm k}_{p2}\bm e_u - \hat{\bm k}_{d2}\bm e_u$ and therefore, we have
\begin{equation*}
\bm v_u^\mathrm{int}=\hat{\bm v}_u^{\mathrm{int}}-\Delta \bm v_u^\mathrm{int},
\end{equation*}
where $\Delta \bm v_u^\mathrm{int} = \Delta \ddot{\bm q}^e_u+\hat{\bm k}_{p2}\Delta \bm q^e_u+ \hat{\bm k}_{d2}\Delta \dot{\bm q}^e_u$. Compared to~\eqref{Eq_BEM}, the BEM estimation error comes from GP modeling error and optimization accuracy. It is reasonable to assume that $\Delta \bm q_u^e$ is bounded. Because of the bounded Gaussian kernel function, the GP prediction variances are also bounded, i.e.,
\begin{equation}
\norm{\bm{\Sigma}_a(\bm{x})} \leq (\sigma^{\max}_{a})^2,
\norm{\bm{\Sigma}_u(\bm{x})} \leq (\sigma^{\max}_{u})^2,
\label{bound}
\end{equation}
where $\sigma^{\max}_{a}=\max _{i}(\sigma_{f_{ai}}^2+\vartheta^2_{ai})^{1/2}$, $\sigma^{\max}_{u}=\max _{i}(\sigma_{f_{ui}}^2+\vartheta^2_{ui})^{1/2}$, $\sigma_{f}$ and $\vartheta$ are the hyperparameters in each channel. Furthermore, we require the control gains to satisfy the following bounds
\begin{align*}
k_{i1}\le \lambda(\hat{\bm k}_{i1})\le k_{i3}, \quad k_{i2} \le  \lambda(\hat{\bm k}_{i2}) \le k_{i4}, \; i=p,d
\end{align*}
for constants $k_{pj},k_{dj}>0$, $j=1,\cdots,4$, where $\lambda(\cdot)$ denotes the eigenvalue operator.

\subsection{PEIC-Based Control Design}

The control design $\bm v^\mathrm{int}$ in~\eqref{Eq_Va_Int} updates the input $\bm v^\mathrm{ext}$ and $\bm q_a$ then acts as a control input to steer $\bm q_u$ to $\bm q_u^e$. The $\mathcal S_u$ dynamics is rewritten into
\begin{equation*}
  \ddot{\bm q}_u=-\bm D_{uu}^{-1}\bm H_u-\bm D_{uu}^{-1}\bm{D}_{u a}\ddot{\bm q}_a.
\end{equation*}
We instead consider the coupling effect between $\bm q_a$ and $\bm q_u$ and assign $m$ control inputs for the unactuated subsystem. To achieve such a goal, we partition the actuated coordinates as $\bm q_a=[\bm q_{aa}^T~\bm q_{au}^T]^T$, $\bm q_{au} \in \mathbb{R}^m$, $\bm q_{aa} \in \mathbb{R}^{n-m}$, and $\bs{u}=[\bs{u}_a^T \; \bs{u}_u^T]^T$. The $\mathcal S^{gp}$ dynamics in~\eqref{Eq_gp_model} is rewritten as
\begin{equation}\label{Eq_dyna1}
\begin{bmatrix}
 \bar{\bm D}_{a a}^a & \bar{\bm D}_{a a}^{au} & \bar{\bm D}_{a u}^a \\
 \bar{\bm D}_{a a}^{ua} &\bar{\bm D}_{a a}^{u} & \bar{\bm D}_{a u}^u\\
 \bar{\bm D}_{u a}^a & \bar{\bm D}_{u a}^u & \bar{\bm D}_{u u}
\end{bmatrix}
\begin{bmatrix}
  \ddot{\bm q}_{aa} \\
  \ddot{\bm q}_{au}\\
  \ddot{\bm q}_u
\end{bmatrix}
+\begin{bmatrix}
  \bm{H}_{aa}^{gp} \\
  \bm{H}_{au}^{gp} \\
  \bm H_{u}^{gp}
\end{bmatrix}
=\begin{bmatrix}
  \bm u_{a} \\
  \bm u_{u} \\
  \bm 0
\end{bmatrix},
\end{equation}
where all block matrices are in proper dimensions. We rewrite~\eqref{Eq_dyna1} into three groups as
\begin{subequations}\label{Eq_S1}
\begin{align}
    \mathcal S^{gp}_{aa}&: \bar{\bm D}_{a a}^a \ddot{\bm q}_{aa}+\bm H_{an}^a=\bm u_{a},\label{Eq_Saa}\\
    \mathcal S^{gp}_{au}&:\bar{\bm D}_{a a}^{u} \ddot{\bm q}_{au}+\bar{\bm D}_{au}^u \ddot{\bm q}_{u}+\bm H_{an}^u=\bm u_{u},\label{Eq_Sau}\\
    \mathcal S^{gp}_{u}&:\bar{\bm D}_{u a}^u \ddot{\bm q}_{au}+\bar{\bm D}_{u u} \ddot{\bm q}_{u}+\bm H_{un}=\bm 0,\label{Eq_Su}
\end{align}
\end{subequations}
where $\bm H_{an}^{a}= \bar{\bm D}_{a a}^{au} \ddot{\bm q}_{au}+ \bar{\bm D}_{a u}^{a} \ddot{\bm q}_u +\bm{H}_{aa}^{gp}$, $\bm H_{an}^u=\bar{\bm D}_{a a}^{ua} \ddot{\bm q}_{aa}+ \bar{\bm D}_{a u}^{u} \ddot{\bm q}_u +\bm{H}_{au}^{gp}$, and $\bm H_{un}=\bar{\bm D}_{u a}^a \ddot{\bm q}_{aa}+ \bm H_u^{gp}$. Apparently, $\mathcal S^{gp}_{u}$ is virtually independent of $\mathcal S^{gp}_{aa}$ and the dynamics coupling exists only between $\mathcal S^{gp}_{u}$ and $\mathcal S^{gp}_{au}$.

Let $\hat{\bm v}^\mathrm{ext}$ in~\eqref{control1} be partitioned into $\hat{\bm v}_{a}^\mathrm{ext}$ and $\hat{\bm v}_{u}^\mathrm{ext}$ corresponding to $\bm q_{aa}$ and $\bm q_{au}$, respectively. $\hat{\bm v}_{a}^\mathrm{ext}$ is directly applied to $\mathcal S^{gp}$ and $\hat{\bm v}_{u}^\mathrm{ext}$ is updated for balance control purpose. As aforementioned, the condition to eliminate the uncontrolled motion in $\mathcal S_a$ is that $\bm q_u$ only depends on $m$ inputs. The task of driving $\bm q_u$ to $\bm q_u^e$ is assigned to $\bm q_{au}$ coordinates only. With this observation, the PEIC-based control takes the form of $\hat{\bm u}^\mathrm{int} =[\hat{\bm u}_{a}^T~\hat{\bm u}_{u}^T]^T$ with
\begin{equation}
\label{Eq_U_PEIC}
    \hat{\bm u}_{a}=\bar{\bm D}_{a a}^a \hat{\bm v}_{a}^{\mathrm{ext}}+\bm H_{a n}^a,\;
    \hat{\bm u}_{u}=\bar{\bm D}_{a a}^u \hat{\bm v}^{\mathrm{int}}+\bar{\bm D}_{a u}^u \ddot{\bm q}_u+\bm H_{a n}^u,
\end{equation}
where $\hat{\bm v}^\mathrm{int} = -\left(\bar{\bm{D}}_{u a}^u\right)^{-1}\left(\bm H_{un}+\bar{\bm D}_{uu} \hat{\bm v}_u^\mathrm{int}\right)$. Clearly, the unactuated subsystem only depends on $\bm u_{u}$ under the PEIC design. The following lemma presents the qualitative assessment of the PEIC-based control and the proof is given in Appendix~\ref{proof_PEIC}.
\begin{lem}
\label{Lem_PEIC}
If conditions $\mathcal C_1$ to $\mathcal C_3$ are satisfied and $\mathcal S^{gp}$ is stable under the EIC-based control design, $\mathcal S^{gp}$ is stable under the PEIC-based control $\hat{\bm u}^\mathrm{int}$.
\end{lem}

\subsection{NEIC-Based Control Design}

Besides the PEIC-based control, we propose an alternative controller in which the control input for $\bm p_{an}$ is explicitly designed. We note that $\bm p_{am}\in\spn(\bm V_m)$ and $\bm p_{an}\in\ker(\bar{\bm D}_{ua})=\spn(\bm V_n)$. The subspace $\spn(\bm V_m)$ and $\spn(\bm V_n)$ are orthogonal in $\mathbb R^{n}$ and the motion of $\bm p_{an}$ is independent of $\bm p_{am}$. Therefore, a compensation is designed in $\spn(\bm V_n)$ for $\bm p_{an}$, which leaves the motion in $\spn(\bm V_m)$ unchanged. Based on this observation, the NEIC-based control is designed as follows.

The NEIC-based control takes the form
\begin{equation}\label{Eq_U_NEIC}
  \tilde{\bm u}^\mathrm{int} =  \bar{\bm D}_{aa} \tilde{\bm v}_{a}^\mathrm{int}+\bar{\bm D}_{a u} \ddot{\bm q}_{u}+\bm{H}_{a}^{gp},
\end{equation}
where $\tilde{\bm v}_{a}^\mathrm{int}= \tilde{\bm v}^\mathrm{int} +\tilde{\bm v}_{an}$, $\tilde{\bm v}_{an}=\bm V_{n}\bm \nu_{n}$, $\tilde{\bm v}^\mathrm{int}=-\bar{\bm{D}}_{u a}^{+}(\bm H_{u}^{gp}+\bar{\bm D}_{uu} \hat{\bm v}_u^\mathrm{int})$, and $\bm \nu_{n}$ is the control design that drives $p_{ai}$ to $p_{ai}^d$, $i=m+1,...,n$. $\bm p_a^d=\bm \Upsilon(\bm q_a^d)$ is transformed reference trajectory. The design of $\bm \nu_{n}$ compensates for the loss of original control effect $\hat{\bm v}^\mathrm{ext}$ to drive $\bm e_a \rightarrow \bm 0$ in $\ker(\bar{\bm D}_{ua})$. A straightforward yet effective design of $\bm \nu_{n}$ can be $\bm \nu_{n}= \alpha\hat{\bm \nu}^\mathrm{ext}_{n}$, where $\alpha>0 $. Compared to the PEIC-based control, $\bm p_{an}$ plays the similar role of $\bm q_{aa}$ coordinates. In the new coordinate, the $\bm q_u$ is associated with $\bm p_{am}$ only.

The following result gives the property of the NEIC-based control and the proof is given in Appendix~\ref{proof_NEIC}.
\begin{lem}
\label{Lem_NEIC}
For $\mathcal S$, if $\mathcal S^{gp}$ satisfies the conditions $\mathcal C_1$ to $\mathcal C_3$ and $\mathcal S^{gp}$ is stable under the EIC-based control,  $\mathcal S^{gp}$ under the NEIC-based control $\tilde{\bm v}_{a}^\mathrm{int}$ is also stable. Meanwhile, $\mathcal S^{gp}_{u}$ is unchanged compared to that under the EIC-based control.
\end{lem}

The proof of Lemma~\ref{Lem_NEIC} show that the inputs $\hat{\bm u}_a^\mathrm{int}$ and $\tilde{\bm u}_a^\mathrm{int}$ follow the control design guidelines. Both the PEIC- and NEIC-based controllers preserve the structured form of the EIC design. Figs.~\ref{Fig_PEIC} and~\ref{Fig_NEIC} illustrate the overall flowchart of the PEIC- and NEIC-based control design, respectively. To take advantages the EIC-based structure, we follow the design guideline to make sure that motion of unactuated coordinates only depends on $m$ inputs in configuration space (PEIC-based control) or transformed space (NEIC-based control). The input $\bm \nu_{n}^\mathrm{ext}$ is re-used for uncontrolled motion under the NEIC-based control. The PEIC-based control assigns the balance task to a partial group of the actuated coordinates.

\section{Control Stability Analysis}
\label{Sec_Stability}

\subsection{Closed-Loop Dynamics Under PEIC-based Control}

To investigate the closed-loop dynamics, we consider the GP prediction error and BEM estimation error. The GP prediction error in~\eqref{modelerror} is extended to $\bs{\Delta}_{aa}$, $\bs{\Delta}_{au}$ and $\bs{\Delta}_u$ for $\bm q_{aa}, \bm q_{au}, \bm q_u$ dynamics, respectively. Under the PEIC-based control, the dynamics of $\mathcal S$ becomes
\begin{align*}
\ddot{\bm q}_{aa}&=\hat{\bm v}_{a}^\mathrm{ext}-(\bar{\bm D}_{aa}^a)^{-1}\bs{\Delta}_{aa},\\
\ddot{\bm q}_{au}&= -(\bar{\bm D}_{u a}^u)^{-1}(\bm H_{u n}+\bar{\bm D}_{uu} \hat{\bm v}_u^\mathrm{int})-(\bar{\bm D}_{a a}^u)^{-1} \bs{\Delta}_{au},\\
\ddot{\bm q}_{u} &=\hat{\bm v}_u^\mathrm{int}-\bar{\bm D}_{uu}^{-1}[\bs{\Delta}_u-\bar{\bm D}_{ua}^u(\bar{\bm D}_{aa}^u)^{-1} \bs{\Delta}_{au}].
\end{align*}
The BEM obtained by~\eqref{Eq_BEM_opt} under input $(\ddot{\bm q}_{aa},\hat{\bm v}_{u}^\mathrm{ext})$ is equivalent to inverting~\eqref{Eq_Su} and therefore, $\hat{\bm v}_{u}^{\mathrm{ext}}=-\left(\bar{\bm D}_{u a}^u\right)^{-1} \bm{H}_{un}\big\vert_{\substack{\bm q_u=\hat{\bm q}_u^{e}, \dot{\bm q}_u=\ddot{\bm q}_u=\bm 0}}$. Substituting the above equation into the $\bm q_{au}$ dynamics yields $\ddot{\bm q}_{au}= \hat{\bm v}_{u}^{\mathrm{ext}}+ \bm O_{au}$, where $\bm O_{au} =-(\bar{\bm D}_{u a}^u)^{-1}\bar{\bm D}_{uu} \hat{\bm v}_u^\mathrm{int}-(\bar{\bm D}_{a a}^u)^{-1} \bs{\Delta}_{au}+\bs{o}_{1}$ and $\bs{o}_1$ denotes the higher order terms.

Defining the total error $\bs{e}_q=[\bs{e}_a^T \; \bs{e}_u^T]^T$ and $\bs{e}=[\bs{e}_q^T \; \dot{\bs{e}}_q^T]^T$, the closed-loop error dynamics becomes
\begin{equation}
\label{Eq_error_dyna1}
\hspace{-0mm} \dot{\bm e}=\underbrace{\begin{bmatrix} \bm 0 & \bs{I}_{n+m} \\ -\hat{\bs{k}}_{p} & -\hat{\bs{k}}_d \end{bmatrix}}_{\bm A} \begin{bmatrix} \bm e_q \\ \dot{\bm e}_q \end{bmatrix}+ \underbrace{\begin{bmatrix} \bm 0 \\ \bs{O}_\mathrm{tot} \end{bmatrix}}_{\bm O_1}=\bm A\bm e + \bm O_1
\end{equation}
with $\bm O_\mathrm{tot} = [\bm O_{a}^T~\bm O_u^T]^T$, $\bm O_a=[\bm O_{aa}^T~\bm O_{au}^T]^T$, $\bm O_{aa} =-(\bar{\bm{D}}_{a a}^a)^{-1}\bs{\Delta}_{aa}$, $ \bm O_u=-\bar{\bm{D}}_{uu}^{-1}(\bs{\Delta}_u-\bar{\bm{D}}_{u a}^u(\bar{\bm{D}}_{a a}^u)^{-1} \bs{\Delta}_{au})-\bs{\Delta} \bm v_u^\mathrm{int}$, $\hat{\bs{k}}_{p}=\diag(\hat{\bs{k}}_{p1},\hat{\bs{k}}_{p2})$, and $\hat{\bs{k}}_{d}=\diag(\hat{\bs{k}}_{d1},\hat{\bs{k}}_{d2})$.

Because of bounded $\bar{\bm D}$, there exists constants $0<d_{a1}, d_{a2}, d_{u1}, d_{u2}<\infty$ such that $d_{a1} \le \norm{\bar{\bm D}_{aa}}\le d_{a2}$ and $d_{u1}\le \norm{\bar{\bm D}_{uu}}\le d_{u2}$. The perturbation terms are further bounded as
\begin{align*}
\norm{\bm{\bm O}_a} = &\norm{-\begin{bmatrix}
\bm 0 \\
(\bar{\bm D}_{u a}^u)^{-1}\bar{\bm D}_{uu} \hat{\bm{v}}_u^{\mathrm{int}}
\end{bmatrix}-
(\bar{\bm D}_{a a}^a)^{-1}\bs{\Delta}_a +\begin{bmatrix} \bm 0 \\ \bm{o}_1
\end{bmatrix}}\nonumber \\
 \leq & \tfrac{d_{u 2}}{\sigma_1}\|\hat{\bm{v}}_u^{\mathrm{int}}\|+\tfrac{1}{d_{a 1}}\norm{\bs{\Delta}_a}+\norm{\bm{o}_1}, \;\; \text{and} \\
\norm{\bm{O}_u}=&\norm{ -\bar{\bm D}_{u u}^{-1}(\bs{\Delta}_u-\bar{\bm D}_{u a}^u(\bar{\bm D}_{a a}^u)^{-1} \bs{\Delta}_{au})-\Delta \bm v_u^{\mathrm{int}}}\nonumber\\
 \leq & \tfrac{1}{d_{u 1}}\norm{\bs{\Delta}_u}+\tfrac{\sigma_m}{d_{u 1} d_{a 1}}\norm{\bs{\Delta}_a}+\|\Delta \bm v_u^\mathrm{int}\|.
\end{align*}
The perturbation $\bm{o}_{1}$ is due to approximation and $\Delta \bs{v}_u^{\mathrm{int}}$ is the control difference by the BEM calculation with the GP prediction. They are both assumed to be affine with $\bs{e}$, i.e.,
\begin{equation}\label{Eq_assumption}
\norm{\bm{o}_1}  \le  {c_1}\norm{\bm e} + c_2,\quad
\|\Delta\bm v_u^{\mathrm {int}}\| \leq c_3\norm{\bm e}+c_4
\end{equation}
with $0 < c_i < \infty$, $i=1,\cdots,4$. From~\eqref{bound}, we have $\|\bm \kappa_a^T \bm \Sigma_a^{\frac{1}{2}} \|\leq \sigma^{\max}_{a}\norm{\bm \kappa_a}$ and $\|\bm \kappa_u^T \bm \Sigma_u^{\frac{1}{2}}\|\leq \sigma^{\max}_{u}\norm{\bm \kappa_u}$. Thus, for $0<\eta=\eta_a \eta_u <1$, we can show that
\begin{equation}\label{Eq_O1}
\Pr\left\{\norm{\bm O_1}\le d_1 +d_2\norm{\bm e}+l_{u1}\norm{\bm \kappa_u}+ l_{a1}\norm{\bm \kappa_a }\right\}\ge \eta,
\end{equation}
where $d_1=c_2+\left(1+\frac{d_{u 2}}{\sigma_1}\right) c_4$, $d_2=c_1+\frac{d_{u 2}}{\sigma_1} c_3$, $l_{a1}=\frac{\sigma^{\max}_{a}(d_{u 1}+\sigma_m)}{d_{u1} d_{a 1}}$, $l_{u1}=\frac{\sigma^{\max}_{u}}{d_{u1}}$.

\subsection{Closed-Loop Dynamics Under NEIC-based Control}

Plugging the NEIC-based control into $\mathcal S^{gp}$ and considering $\bm \Delta$, we obtain
\begin{subequations}
\label{Eq_S_NEIC2}
\begin{align}
\ddot {\bm p}_{am}&=-\bm \Lambda_m^{-1}\bm U^T(\bm H_u^{gp}+\bar{\bm D}_{uu}  \hat{\bm v}_u^{\mathrm{int}})-\bm \Lambda_m^{-1}\bm U^T\bm \Delta_{u} -\bm V_{m}^T\bar{\bm D}_{aa}^{-1}\bm \Delta_a, \label{Eq_pam_NEIC}\\
\ddot{\bm p}_{an} &=\bm \nu_{n}^\mathrm{ext}-{\bm V_{n}^T\bar{\bm D}_{aa}^{-1}\bm \Delta_a},\label{Eq_pan_NEIC}
\\
\ddot{\bm q}_{u} &= \hat{\bm v}_{u}^\mathrm{int}-\bar{\bm D}_{uu}^{-1}(\bm \Delta_u- \bar{\bm D}_{ua}\bar{\bm D}_{aa}^{-1}\bm \Delta_a).
\end{align}
\end{subequations}
To obtain the error dynamics, we take advantage of the definition of BEM and from~\eqref{Eq_Gamma_BEM_Vu}, we have
$\bm \nu_{a}^{\mathrm{ext}}=-\bm \Lambda_m^{-1} \bm{U}^T \bm{H}_{u}^{gp}\big\vert_{\bm q_u=\hat{\bm q}_u^{e}, \dot{\bm q}_u=\ddot{\bm q}_u=\bm 0}$. Then we rewrite~\eqref{Eq_pam_NEIC} into
\begin{align}\label{Eq_pam_approx}
\ddot {\bm p}_{am} &=-\bm \Lambda_m^{-1} \bm{U}^T \bm{H}_{u}^{gp}\big\vert_{\substack{{\bm q}_u=\hat{\bm q}_u^{e}\\
\dot{\bm q}_u=\ddot{\bm q}_u=\bm 0}} +\bm o_2-\bm \Lambda_m^{-1} \bm{U}^T \bar{\bm D}_{u u} \hat{\bm v}_u^\mathrm{int}\nonumber\\
&\quad -\bm \Lambda_m^{-1}\bm U^T\bm \Delta_{u} -\bm V_{m}^T\bar{\bm D}_{aa}^{-1}\bm \Delta_a=\bm \nu_{m}^\mathrm{ext} +  \bm O_{m},
\end{align}
where $\bm o_2$ is the residual that contains higher order terms. $\bm O_{am}=\bm o_2-\bm \Lambda_m^{-1} \bm{U}^T \bar{\bm D}_{u u} \hat{\bm v}_u^\mathrm{int} -\bm \Lambda_m^{-1}\bm U^T\bm \Delta_{u} -\bm V_{m}^T\bar{\bm D}_{aa}^{-1}\bm \Delta_a$ denotes the total perturbations.

The $\mathcal S_u^{gp}$ dynamics keeps the same form as that in the PEIC-based control. We write the error dynamics under the NEIC-based control as
\begin{subequations}
\begin{align}
&\ddot {\bm e}_{am} =-\hat{\bm k}_{p1}\bm e_{am}-\hat{\bm k}_{d1}\bm e_{am}+ \bm O_{am},\label{Eq_err_qan_dynb}\\
&\ddot {\bm e}_{an} =-\hat{\bm k}_{p1}\bm e_{an}-\hat{\bm k}_{d1}\bm e_{an}+ \bm O_{an},\label{Eq_err_qan_dyna}\\
&\ddot{\bm e}_u =-\hat{\bm k}_{p2}\bm e_u-\hat{\bm k}_{d2}\bm e_u+ \bm O_{u},
\end{align}
\end{subequations}
\hspace{-0.1mm}where $\bm e_{am} = \bm p_{am}-\bm p_{am}^d$, $\bm e_{an} = \bm p_{an}-\bm p_{an}^d$, $\bm p_a^d=\bs{V}^T\bm q_a^d$ is the image of $\bm q_a^d$ under $\bm\Upsilon$, and $\bm O_{an}=-{\bm V_{n}^T\bar{\bm D}_{aa}^{-1}\bm \Delta_a}$. Applying inverse mapping $\bm\Upsilon^{-1}$ to~\eqref{Eq_err_qan_dynb} and~\eqref{Eq_err_qan_dyna}, the error dynamics in $\bs{q}$ is obtained as
\begin{equation}
\mathcal{S}_{e,\mathrm{NEIC}}: \; \dot{\bm e}=  \bm A\bm e + \bm O_2,
\label{errordynNEIC}
\end{equation}
where $\bm O_2$ is the transformed perturbations of $[\bm O_{an}^T~\bm O_{am}^T~\bm O_{u}^T]^T$. Following the same steps in the above analysis of PEIC-based control, we obtain
\begin{equation}\label{Eq_O2}
\Pr\left\{\norm{\bm O_2}\le d_1 +d_2\norm{\bm e}+l_{u2}\norm{\bm \kappa_u}+l_{a2}\norm{\bm \kappa_a}\right\}\ge \eta,
\end{equation}
where $l_{u2} = \sigma_{u,\max}\frac{\sigma_1+d_{u 1}}{\sigma_1 d_{u 1}}$, and $l_{a2 }=\sigma_{a,\max}\frac{\sigma_m+d_{u 1}}{d_{a 1} d_{u 1}}$. The assumption in~\eqref{Eq_assumption} is also used for $\bm o_2$.

\subsection{Stability Results}

To show the stability, we consider the Lyapunov function candidate $V=\bm e^T \bm P\bm e\ge0$, where positive definite matrix $\bm P=\bm P^T$ is the solution of
\begin{equation}\label{Eq_Q}
  \bm A_0^T \bm P +\bm P \bm A_0+\bm Q=\bm 0, \;\;  \bs{A}_0=\begin{bmatrix} \bm 0 & \bs{I}_{n+m} \\ -\bs{k}_{p} & -\bs{k}_d \end{bmatrix}
\end{equation}
for given positive definite matrix $\bm Q=\bm Q^T$, where $\bm  A_0$ is the constant part of $\bs{A}$ in~\eqref{Eq_error_dyna1} and does not depend on variances $\bs{\Sigma}_a$ or $\bs{\Sigma}_u$. $\bs{k}_{p}=\diag(\bs{k}_{p1},\bs{k}_{p2})$ and $\bs{k}_{d}=\diag(\bs{k}_{d1},\bs{k}_{d2})$.

We denote the corresponding Lyapunov function candidates for the NEIC- and PEIC-based controls as $V_1$ and $V_2$, respectively. The stability results are summarized as follows with the proof is given in Appendix~\ref{proof_error_bound}.
\begin{thm}
\label{Thm_Error_Bound}
For robot dynamics~\eqref{Eq_Sub_Dynamics}, using the GP-based model~\eqref{Eq_gp_model} that satisfies conditions $\mathcal C_1$-$\mathcal C_3$, under the PEIC- and NEIC-based control, the Lyapunov function under each controller satisfies
\begin{equation}
  \mathrm{Pr} \left\{\dot V_i\le -\gamma_i V_i + \rho_i + \varpi_i\right\} \geq \eta, \; i=1,2,
\end{equation}
and the error $\bs{e}$ converges to a small ball around the origin, where $\gamma_i$ is the convergence rate, $\rho_i$ and $\varpi_i$ are the perturbation terms, and $0<\eta=\eta_a\eta_u<1$.
\end{thm}


\setcounter{figure}{2}
\begin{figure*}[th!]
	\centering
	\subfigure[]{
		\label{Fig_pen_arm}
		\includegraphics[height=4.0cm]{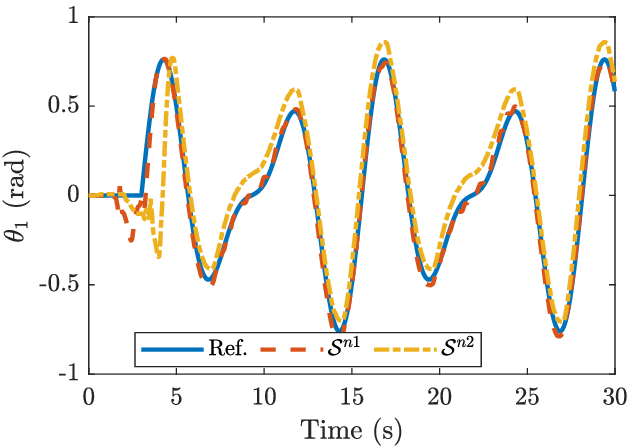}}
	\hspace{-3.0mm}
	\subfigure[]{
		\label{Fig_pen_pen}
		\includegraphics[height=4.0cm]{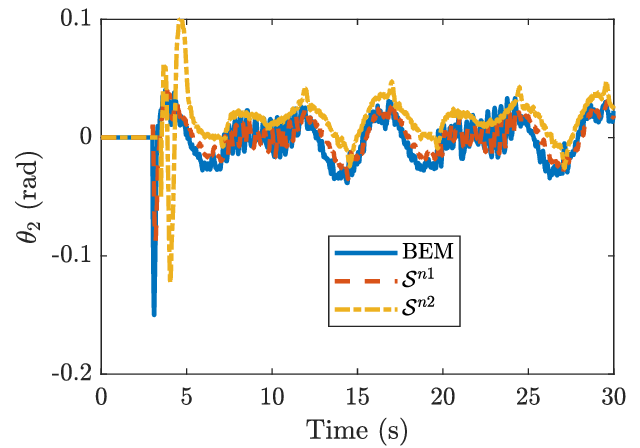}}
	\hspace{-3mm}
	\subfigure[]{
		\label{Fig_pen_error}
		\includegraphics[height=4.0cm]{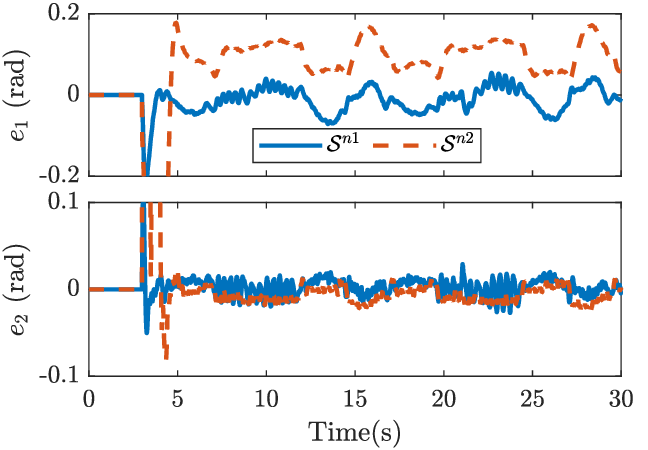}}
	\subfigure[]{
		\label{Fig_Profile_pen}
		\includegraphics[height=4.1cm]{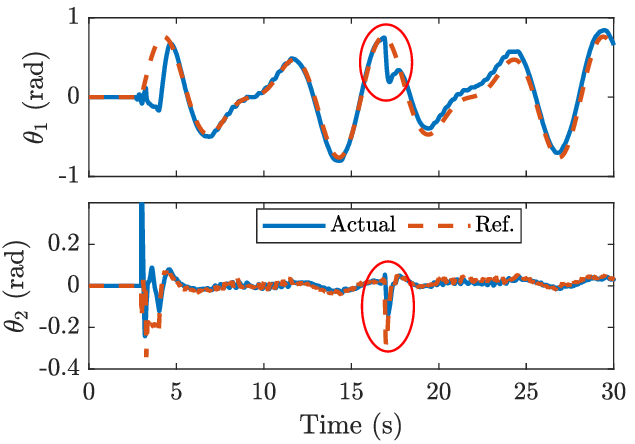}}
	\hspace{-4.5mm}
	\subfigure[]{
		\label{Fig_V_pen}
		\includegraphics[height=4.1cm]{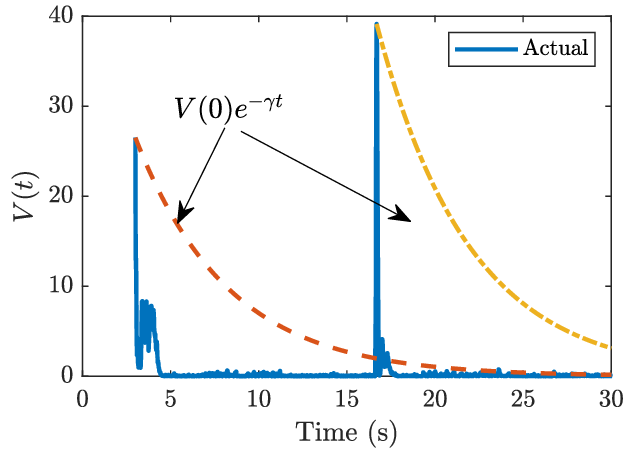}}
	\hspace{-5mm}
	\subfigure[]{
		\label{Fig_Error_Bound_pen}
		\includegraphics[height=4.cm,width=2.4in]{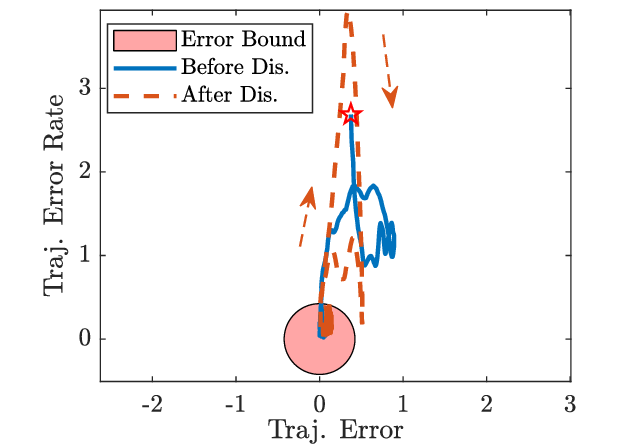}}
	\vspace{-4mm}
	\caption{Experiment results with guaranteed performance.  (a) Arm rotation angle,  (b) pendulum rotation angle, and (c) tracking control error under GP-based control.  (d) Pendulum motion profile.  (e) Profile of Lyapunov function. (f) Trajectory error motion. At $t=17$~s, an impact disturbance is applied. The dashed arrow in (f) indicates the direction in which the error grows after disturbance is applied.}
	\label{Fig_pendulum}
	\vspace{-2mm}
\end{figure*}

\vspace{-2mm}	
\section{Experimental Results}
\label{Sec_Result}

Two inverted pendulum platforms are used to conduct experiments to validate the control design. The results from each platform demonstrate different aspects of the control performance.

\vspace{-2mm}	
\subsection{2-DOF Rotary Inverted Pendulum}

Fig.~\ref{Fig_pen_photo} shows a 2-DOF rotary inverted pendulum that was fabricated by Quanser Inc. The base joint ($\theta_1$) is actuated by a DC motor and the inverted pendulum joint ($\theta_2$) is unactuated, and therefore $n=m=1$. We use this platform to illustrate the EIC-based control and also compare the performance under different models and controllers. The robot physical model is given in~\cite{Apk2011} and can be found in Appendix~\ref{Append_Rotary}.

\setcounter{figure}{1}
\vspace{-0mm}	
\begin{figure}[h!]
	\centering
	\subfigure[]{
		\label{Fig_pen_photo}
		\includegraphics[height=4.3cm]{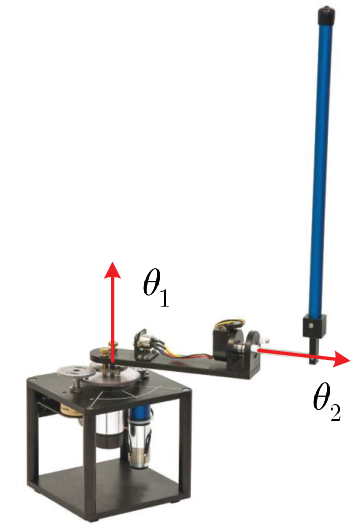}}
	\hspace{5mm}
	\subfigure[]{
		\label{Fig_Leg_photo}
		\includegraphics[height=4.2cm]{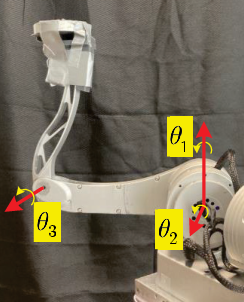}}
	\vspace{-1mm}	
	\caption{(a) A Furuta pendulum. The base link joint $\theta_1$ is actuated and the pendulum link joint  $\theta_2$ is unactuated. (b) A 3-link inverted pendulum with actuated joints $\theta_1$ and $\theta_2$ and unactuated joint $\theta_3$.}
	\vspace{-2mm}
\end{figure}

\setcounter{figure}{3}

Because $m=n=1$ and the unactuated coordinate depends on the control input, there is no uncontrolled motion when the EIC-based control is applied. Therefore, either a constant or time-varying nominal model would work for the GP-based learning control. We created the following two nominal models
\begin{align*}
\mathcal S^{n_1}:& \;\; \bar{\bm D}_1=\frac{1}{100}\begin{bmatrix}
		5 & -2\c_{2}\\
		-2\c_{2} & 2
	\end{bmatrix},~\bar{\bm H}_1=\begin{bmatrix}
		0 \\
		-\s_2
	\end{bmatrix}, \\
\mathcal S^{n_2}: & \;\; \bar{\bm D}_2=\frac{1}{100}\begin{bmatrix}
		2 & 1\\
		1 & 2
	\end{bmatrix},~\bar{\bm H}_2=\bm 0,
\end{align*}
where $\c_i=\cos\theta_i$, $\s_i=\sin\theta_i$ for angle $\theta_i$, $i=1,2$. The training data were sampled and obtained from the robot system by applying control input $u=\bm k^T[\theta_1- \theta_1^t\;\;\theta_2 \;\;\dot{\theta}_1-\dot{\theta}_1^t \; \;\dot \theta_2]^T$, where $\bm k \in \mathbb R^{1\times 4}$ and $\theta_1^t$ was the combination of sine waves with different amplitudes and frequencies. We chose this input to excite the system and the gain $\bm k$ was selected without the need to balance the platform.

We trained the GP regression models using a total of $500$ data points that were randomly selected from a large dataset. We designed the control gains as $\hat{k}_{p1}=10+50\Sigma_a$, $\hat{k}_{d1}=3+10\Sigma_a$, $\hat{k}_{p2}=1000+500\Sigma_u$, and $\hat{k}_{d2}=100+200\Sigma_u$. The variances $\Sigma_a$ and $\Sigma_u$ were updated online with new measurements in real time. The reference trajectory was $\theta^d_1=0.5\sin t+0.3\sin 1.5t $~rad. The control was implemented at $400$~Hz in Matlab/Simulink real-time system.

Figs.~\ref{Fig_pen_arm} and~\ref{Fig_pen_pen} show the tracking of $\theta_1$ and balance of $\theta_2$ under the EIC-based control. With either $\mathcal S^{n_1}$ or $\mathcal S^{n_2}$, the base link joint $\theta_1$ closely followed the reference trajectory $\theta_1^d$ and the pendulum link joint $\theta_2$ was stabilized around its equilibrium  $\theta^e_2$ as well. The tracking error was reduced further and the pendulum closely followed the small variation under $S^{n_1}$. With $\mathcal S^{n_2}$, the tracking errors became large when the base link changed rotation direction; see Fig.~\ref{Fig_pen_error} at $t=10, 17, 22$~s. Both the time-varying and constant nominal models worked for the EIC-based learning control.

Table~\ref{Table_pen_eror} further lists the tracking errors (mean and one standard deviation) under the both GP models. For comparison purposes, we also conducted additional experiments to implement the physical model-based EIC control and the GP-based MPC design in~\cite{ChenTRO2022}. The tracking and balance errors under the EIC-based learning control with model $S^{n_1}$ are the smallest. In particular, with the time-varying model $S^{n_1}$, the mean values of tracking errors ${e}_1$ and $e_2$ were reduced by $75\%$ and $65\%$ respectively in comparison with those under the physical model-based EIC control. Compared with the MPC method in~\cite{ChenTRO2022}, the tracking errors with nominal model $\mathcal S^{n_2}$ are at the same level.

\vspace{-1mm}
\renewcommand{\arraystretch}{1.4}
\setlength{\tabcolsep}{0.04in}
\begin{table}[h!]
	\centering
\vspace{-2mm}
	\caption{Tracking errors comparison under various controllers ($\times 10^{-1}$ rad)}
	\vspace{2mm}
	\label{Table_pen_eror}
	{\small  \begin{tabular}{|c|c|c|c|c|}
			\hline\hline
			& $\mathcal S^{n_1}$ & $\mathcal S^{n_2}$ & GP-based MPC~\cite{ChenTRO2022} & Physical EIC \\ \hline
			$|e_1|$& $0.24\pm 0.17$ & $0.96\pm 0.34$ & $0.87\pm 0.52$ & $1.09\pm 0.40$ \\ \hline
			$|e_2|$& $0.09\pm0.05$ & $0.09\pm 0.39$ & $0.07\pm 0.06$ & $0.26\pm 0.15$ \\
			\hline\hline
	\end{tabular}}
	\vspace{-2mm}
\end{table}


\begin{figure*}
	\centering
	\subfigure[]{
		\label{Fig_Leg_Traj_PEIC}
		\includegraphics[height=4.75cm]{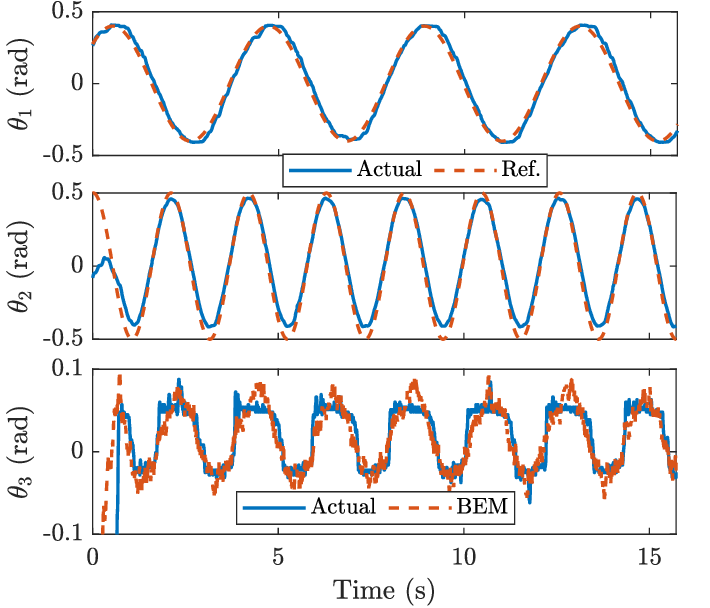}}
	\hspace{-2mm}
	\subfigure[]{
		\label{Fig_Leg_Traj_NEIC}
		\includegraphics[height=4.75cm]{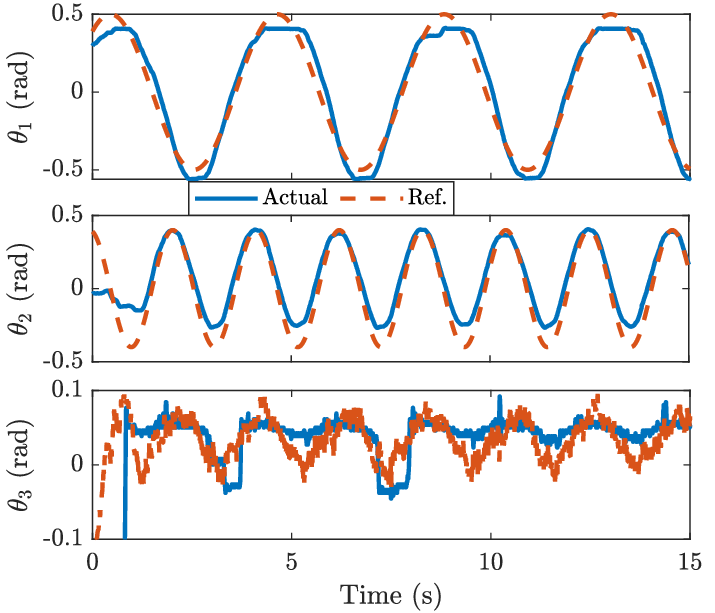}}
	\hspace{-2mm}
	\subfigure[]{
		\label{Fig_Leg_Error_PEIC}
		\includegraphics[height=4.75cm]{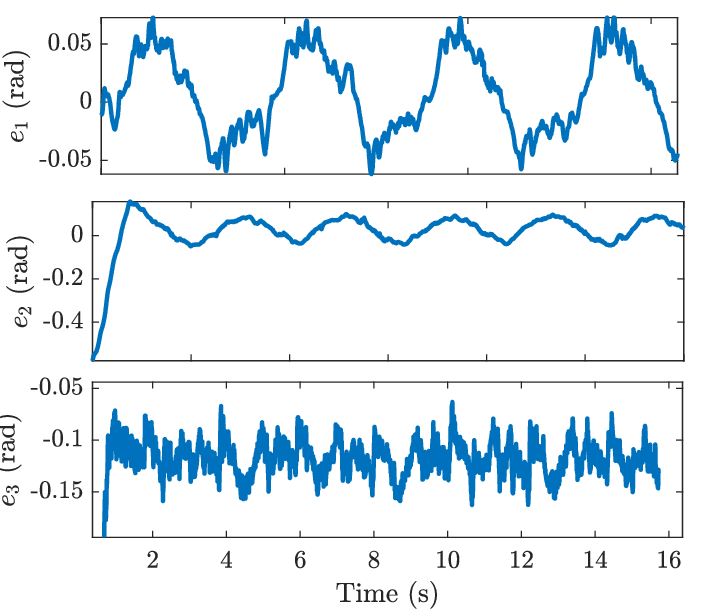}}
	\hspace{-2mm}
	\subfigure[]{
		\label{Fig_Leg_Error_NEIC}
		\includegraphics[height=4.75cm]{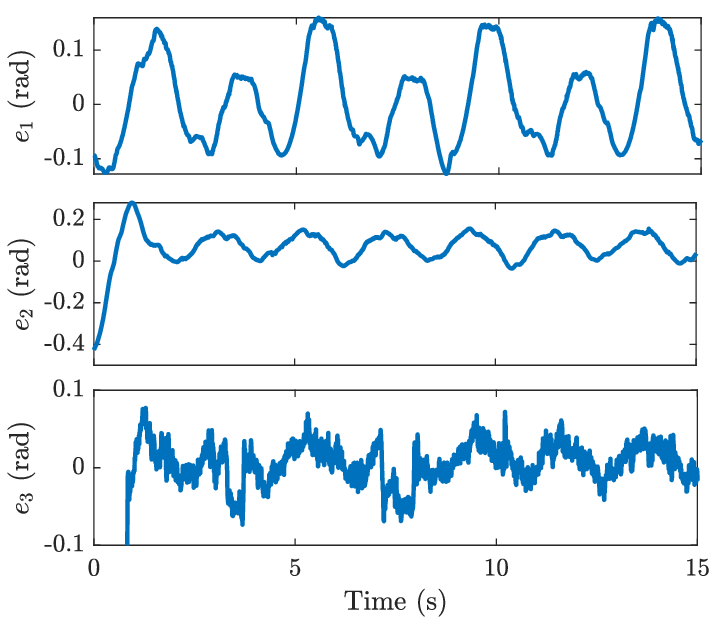}}
	\hspace{-2mm}
	\subfigure[]{
		\label{Fig_Leg_Error_Bound}
		\includegraphics[height=4.72cm]{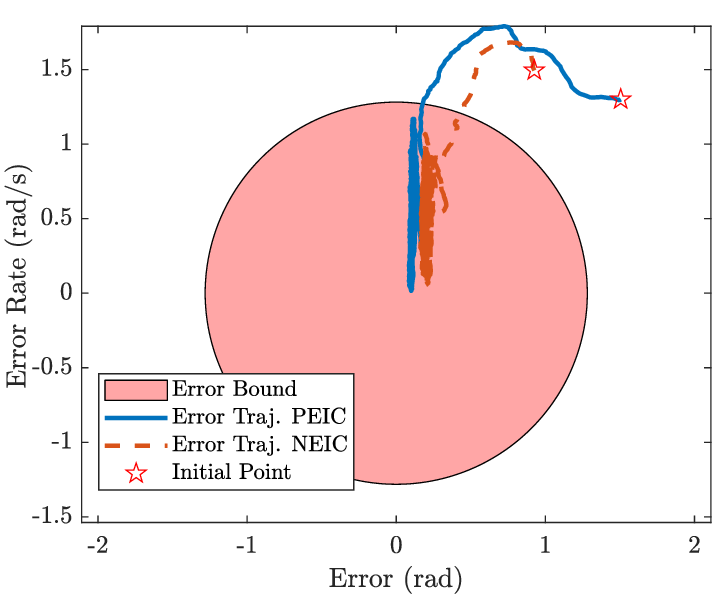}}
	\hspace{-2mm}
	\subfigure[]{
		\label{Fig_Leg_V}
		\includegraphics[height=4.7cm]{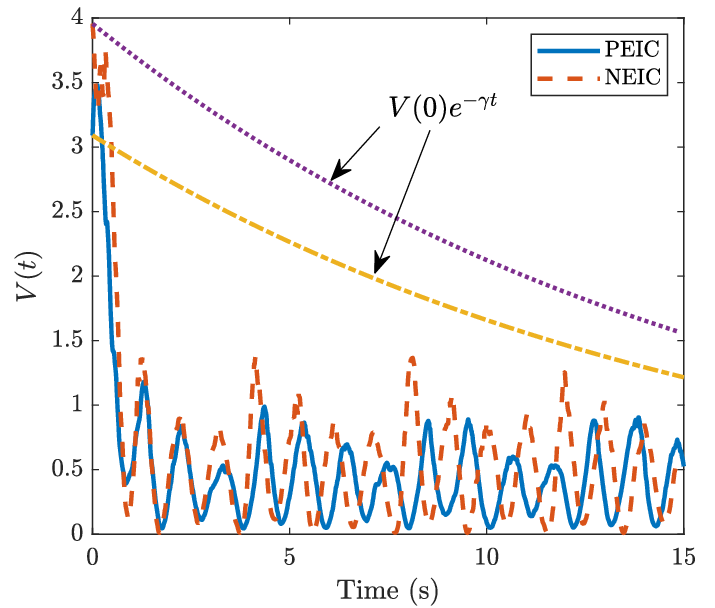}}
	\vspace{-1mm}		
	\caption{Experiment results with the 3-DOF inverted pendulum. (a) and (b) Motion profiles under the PEIC- and NEIC-based control. (c) and (d) Tracking errors under the PEIC- and NEIC-based control. (e) Error trajectory in the $\|\bm e_q\|$-$\|\dot{\bm e}_q\|$ plane. (f) Comparison of the estimated Lyapunov function profile with the actual one.}
	\label{Fig_Leg}
	\vspace{-2mm}	
\end{figure*}

Fig.~\ref{Fig_Profile_pen} shows that the control performance with nominal model $\mathcal S^{n_1}$ under disturbance. At $t=17$~s, an impact disturbance (by manually pushing the pendulum link) was applied and the joint angles changed rapidly with $\Delta \theta_1=0.7$~rad and $\Delta \theta_2=0.3$~rad. The control gains increased ($\hat{k}_{p2}=1215$, $\hat{k}_{d2}=143$) to respond to the disturbance. As a result, the pendulum motion tracked the BEM closely and maintained the pendulum balance after the impact disturbance. Fig.~\ref{Fig_V_pen} shows the calculated Lyapunov function candidate $V(t)$ and its envelope (i.e., $V(t)=V(0)e^{-\gamma t}$, $\gamma=0.1898$) during the experiment. Fig.~\ref{Fig_Error_Bound_pen} shows the error trajectory in the $\|{\bm e}_q\|$-$\|\dot {\bm e}_q\|$ plane. The solid/dashed line shows the error trajectory before/after impact disturbance. The tracking error converged quickly into the error bound. After the disturbance was applied at $t=17$~s, both the Lyapunov function and errors grew dramatically. As the control gains increased, the errors  quickly converged back to the estimated bound again.



\begin{figure*}
	\centering
	\subfigure[]{
		\label{Fig_Leg_PEIC_P}
		\includegraphics[height=4cm]{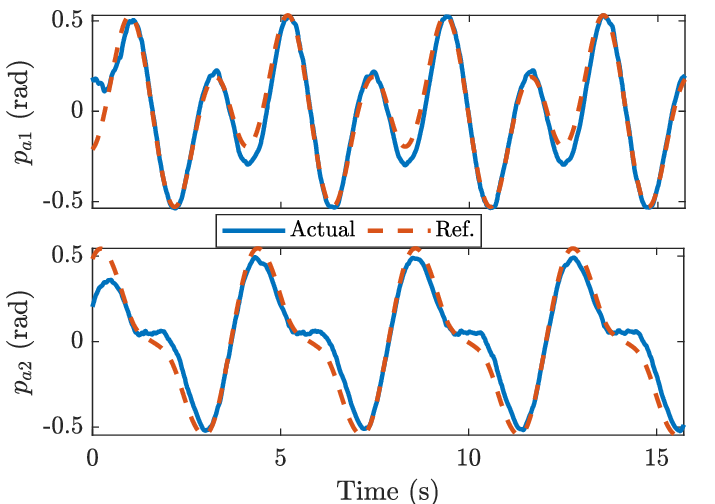}}
	\hspace{-4mm}
	\subfigure[]{
		\label{Fig_Leg_NEIC_P}
		\includegraphics[height=4cm]{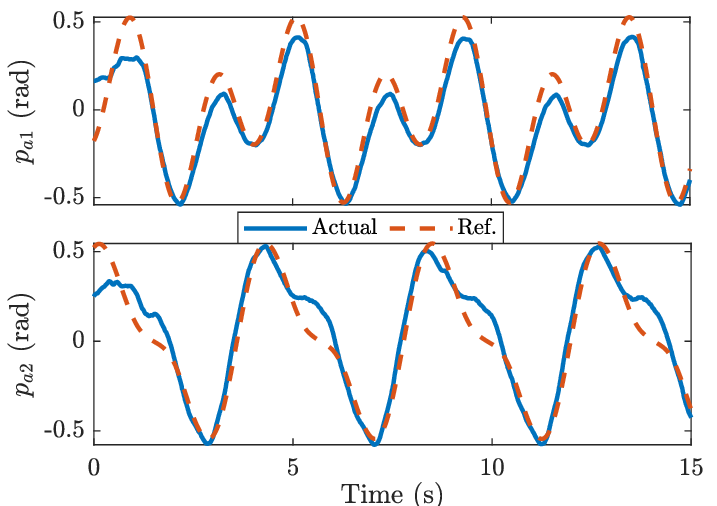}}
	\hspace{-4mm}
	\subfigure[]{
		\label{Fig_Leg_EIC_P}
		\includegraphics[height=4cm]{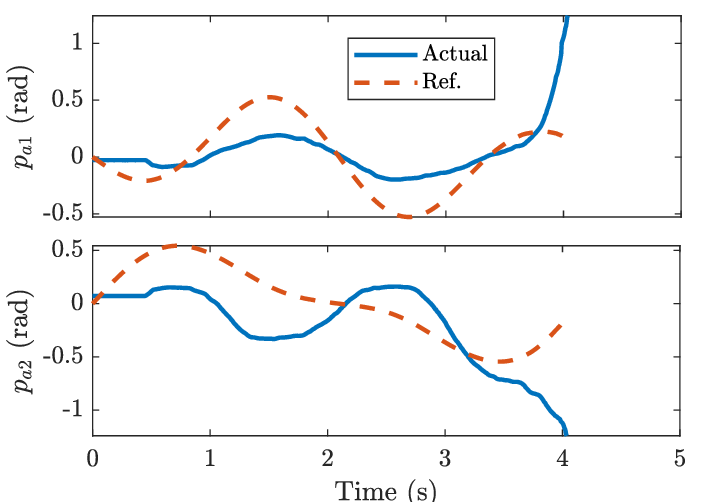}}
\vspace{-2mm}	
\caption{Motion profiles of robotic leg in transformed space under (a) PEIC-based control, (b) NEIC-based control, and (c) EIC-based control.}
	\label{Fig_Leg_pa}
\vspace{-0mm}
\end{figure*}

\vspace{-3mm}
\subsection{3-DOF Rotary Inverted Pendulum}

Fig.~\ref{Fig_Leg_photo} for a 3-DOF inverted pendulum with two actuated joints ($\theta_1$ and $\theta_2$) and one unactuated joint ($\theta_3$), namely, $n=2, m=1$. The physical model of the robot dynamics was obtained using the Lagrangian method and is given in Appendix~\ref{Append_Leg}. All controllers were implemented at an updating frequency of $200$~Hz through the Robot Operating System (ROS). Given the results in the previous example, the time-varying nominal model was selected as
\begin{equation*}
  \bar{\bm D} = \begin{bmatrix}
        0.15 & 0.025\c_2 & 0.025\c_3 \\
        0.025\c_2 & 0.15 & 0.05\c_{2-3} \\
        0.025\c_3 & 0.05\c_{2-3} & 0.1
      \end{bmatrix},~
  \bar{\bm H}=\begin{bmatrix}
      0 \\
      0.2\c_2\\
      0.1\s_3
    \end{bmatrix},
\end{equation*}
where $\c_{i\pm j}=\cos(\theta_i\pm \theta_j)$. We applied an open-loop control (combination of sine wave torque) to excite the system and obtain the training dataset for the three GP models (i.e., one for each joint). The control gains were $\hat{\bs{k}}_{p1} = 15\bm I_2+20\bm \Sigma_a, \hat{\bs{k}}_{d1} = 3\bm I_2+10\bm \Sigma_a, \hat{k}_{p2} = 25+20\Sigma_u, \hat{k}_{d2} = 5.5+10\Sigma_u$, where GP variances $\bm \Sigma_a$ and $\Sigma_u$ were updated online in real time. The reference trajectory was chosen as $\theta_1^d=0.5\sin 1.5t$, $\theta_2^d=0.4\sin 3t$~rad.

\renewcommand{\arraystretch}{1.4}
\begin{table*}[ht]
\caption{Statistical Analysis of Tracking Performance (Mean and Standard Deviation for Errors) Under Different Controllers}\label{Tab_Leg_Error}
\vspace{2mm}
\centering
\setlength{\tabcolsep}{0.1in}
{\small
\begin{tabular}{|c|c|c|c|c|c|}
\hline\hline & $|e_1|$~(rad) & $|e_2|$~(rad) & $|e_3|$~(rad) & $\norm{\bm e}$&$\int \bm u^T \bm u dt$\\
\hline PEIC (GP) & $0.0302\pm0.0178$ & $0.0566\pm0.0685$ & $0.1182\pm 0.0160$ & $0.1343\pm0.0166$ & $5.7659$\\
\hline NEIC (GP, $\alpha=0.5$) & $0.1395\pm 0.0946$ & $0.1166\pm 0.0512$ & $0.0303\pm 0.0209$ & $0.2001\pm 0.0770$ & $5.9022$\\
\hline NEIC (GP, $\alpha=1.0$) & $ 0.0651\pm 0.0416$ & $0.0756\pm 0.0481$ & $0.0195\pm 0.0152$ & $0.1101\pm 0.0499$ & $5.7089$\\
\hline NEIC (GP, $\alpha=1.5$) & $0.0376\pm0.0302$  & $0.0792\pm0.0482$ & $0.0207\pm0.0169$ & $0.0972\pm0.0470$ & $5.7305$  \\
\hline  PEIC (Model) & $0.2168\pm0.1165$  & $0.2398\pm0.1649$ & $0.0179\pm0.0140$ & $0.3587\pm0.1307$ & $5.7978$ \\
\hline NEIC (Model, $\alpha=1.0$) & $0.1374\pm0.0922$  & $0.1237\pm0.0597$ & $0.0455\pm0.0385$ & $0.2095\pm0.0769$ & 5.8452\\
\hline\hline
\end{tabular}}
\vspace{-1mm}
\end{table*}

For the PEIC-based control, we chose $q_{aa}=\theta_1$ and $q_{au}=\theta_2$ and the NEIC-based control was $\bm \nu_{n}=\hat{\bm \nu}_{n}^\mathrm{ext}$. Fig.~\ref{Fig_Leg} shows the experimental results under the PEIC- and NEIC-based control. Under both controllers, the actuated joints ($\theta_1$ and $\theta_2$) followed the given reference trajectories ($\theta_1^d$ and $\theta_2^d$) closely and the unactuated joint ($\theta_3$) was balanced around the BEM ($\theta_3^e$) as shown in Figs.~\ref{Fig_Leg_Traj_PEIC} and~\ref{Fig_Leg_Traj_NEIC}. The pendulum link motion displayed a similar pattern for both controllers. However, the tracking error $e_1$ under the PEIC-based control (i.e., from $-0.05$ to $0.05$~rad) was much smaller than that under the NEIC-based control (i.e., from $-0.15$ to $0.15$~rad); see Figs.~\ref{Fig_Leg_Error_PEIC} and~\ref{Fig_Leg_Error_NEIC}. The balance task in the PEIC-based control was assigned to joint $\theta_2$ and joint $\theta_1$ is viewed as virtually independent of $\theta_2$ and $\theta_3$. Joint $\theta_1$ achieved almost-perfect tracking control regardless of the errors for $\theta_2$ and $\theta_3$. The compensation effect in the null space appeared in the entire configuration space and any motion error in the unactuated joints affected the motion of all actuated joints. Similar to the previous example, Fig.~\ref{Fig_Leg_Error_Bound} shows the error trajectory profile in the $\|\bs{e}_q\|$-$\|\dot{\bs{e}_q}\|$ plane. Fig.~\ref{Fig_Leg_V} shows the Lyapunov function profiles under the PEIC- and NEIC-based controls.

Fig.~\ref{Fig_Leg_pa} shows the motion of the actuated coordinate in the transformed coordinate $\bs{p}_a$ under various controllers. Under the PEIC- and NEIC-based controls, the $\bm p_a$ variables followed the reference profile $\bm p_a^d$ as shown in Figs.~\ref{Fig_Leg_PEIC_P} and~\ref{Fig_Leg_NEIC_P}. Fig.~\ref{Fig_Leg_EIC_P} shows the motion profile under the EIC-based control. In the first 2 seconds, joint $\theta_3$ followed the BEM under the EIC-based control and $p_{a1}$ coordinates displayed a similar motion pattern. However, $p_{a2}$ coordinate showed diverge behavior and led to fall completely. Therefore, as analyzed previously, the system became unstable under the EIC-based control though conditions $\mathcal C_1$ to $\mathcal C_3$ were satisfied.

\begin{figure}[h!]
\centering
\includegraphics[width=8.2cm]{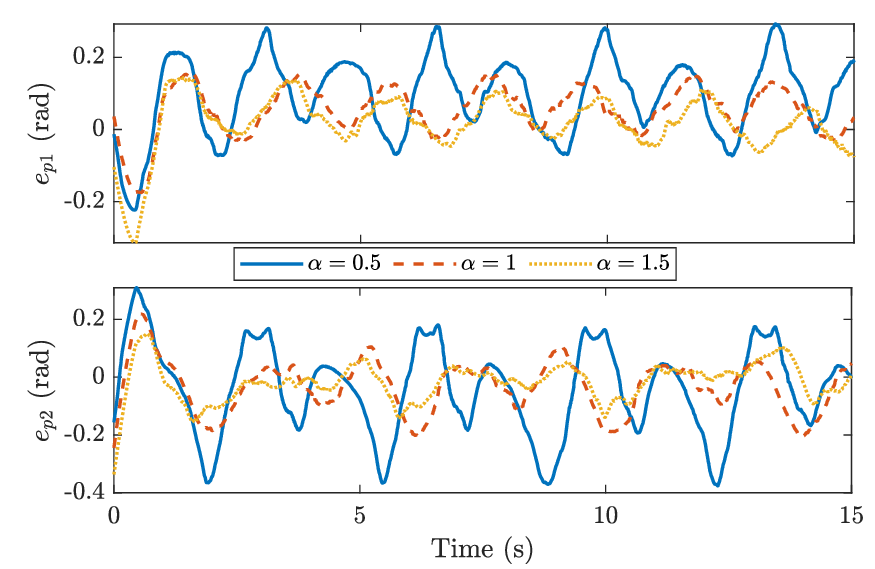}
  \vspace{-2mm}
  \caption{The tracking errors in coordinate $\bs{p}_a$ under the NEIC-based control with various $\alpha$ values.}
  \label{Fig_leg_error_alpha}
  \vspace{-3mm}
\end{figure}

For the fixed NEIC-based control $\bm v^\mathrm{int}$ for balancing the unactuated subsystem, we can reduce the tracking error by updating $\bm \nu_n$ with increased $\alpha$ values. Fig.~\ref{Fig_leg_error_alpha} shows the experiment results of the $\bm p_a$ error profiles under various $\alpha$ values varying from $0.5$ to $1.5$. With a large $\alpha$ value, the tracking error of actuated coordinates was reduced. Table~\ref{Tab_Leg_Error} further lists the steady-state errors (in joint angles) under the NEIC-based control with various $\alpha$ values, the PEIC-based control and the physical model-based control design. Under the NEIC-based control with $\alpha =0.5$, the system was stabilized; when increasing $\alpha$ values to $1$ and $1.5$, the mean tracking errors were reduced $50\%$ and $70\%$ for $\theta_1$, respectively, and $40\%$ for $\theta_2$. Since control input $\bm \nu_n$ did not affect the balance task of the unactuated subsystem, the tracking errors for $\theta_3$ maintained at the same level. It is of interest that the control effort (i.e., last column in Table~\ref{Tab_Leg_Error}) only shows a slight increase with large $\alpha$ values.

\subsection{Discussion}

For the rotary pendulum example, we have $n=m$ and the null space $\ker(\bm D_{au})$ vanishes. The compensation effect is no longer needed by the NEIC-based control, i.e., $\tilde{\bm v}_{a}^\mathrm{int}= \tilde{\bm v}^\mathrm{int}$ and $\tilde{\bm u}^\mathrm{int} =  \bar{\bm D}_{aa} \tilde{\bm v}_{a}^\mathrm{int}+\bar{\bm D}_{a u} \ddot{\bm q}_{u}+\bm{H}_{a}^{gp}={\bm u}^\mathrm{int}$. In this case, the PEIC- and NEIC-based controls are degenerated to the EIC-based control. For the 3-DOF inverted pendulum, the control inputs $u_1$ and $u_2$ act on $\theta_3$ joints through $\ddot{\theta}_1$ and $\ddot{\theta}_2$. Therefore, as shown in Lemma~\ref{Lem_Uncontrolled_Motion}, the uncontrolled motion exists since all controls show up in $\mathcal S_u$ dynamics. This observation explains why the EIC-based control failed to balance the three-link inverted pendulum. If the $\mathcal S_u$ dynamics is related to $m$ control inputs (through $\ddot{\bs{q}}_a$) for $n>m$ such as the bikebot dynamics in~\cite{ChenTRO2022,WangJDSMC2023}, only $m$ external controls were updated and the EIC-based control worked well without any uncontrolled motion.

For the PEIC-based control, the robot dynamics were partitioned into $\mathcal{S}^{gp}=\left\{\mathcal S_{aa}^{gp}, \{\mathcal S_{au}^{gp},\mathcal S^{gp}_{u}\}\right\}$, which contains a fully actuated system $\mathcal S_{aa}^{gp}$, and a reduced-order underactuated system $\{\mathcal S_{au}^{gp}, \mathcal S^{gp}_{u}\}$. The EIC-based control is applied to $\mathcal S_{au}^{gp}$ and $\mathcal S^{gp}_{u}$ only.  The dynamics of $\bm q_u$ in general does not depend on any specific $m$ actuated coordinates, since the mapping $\bm \Upsilon$ is time-varying across different control cycles. In the NEIC-based control design, $\bm p_{am}$ and $\bm q_u$ become an underactuated subsystem and $\bm p_{an}$ is fully actuated.

In practice, no specific rules are specified to select $\bm q_{au}$ out of $\bm q_a$ coordinates, and therefore, there are a total of $C_{n}^{m}=\frac{n!}{m!(n-m)!}$ options to select different coordinates. We can take advantage of such a property to optimize tracking performance for selected coordinates. In the 3-DOF pendulum case, we assigned the balance task of $\theta_3$ to $\theta_2$ motion. The length of link 1 was only $0.09$~m, which was much shorter than the length of link 2 ($0.23$~m). The coupling effect between $\theta_2$ and $\theta_3$ was much stronger than that between $\theta_1$ and $\theta_3$; see $D_{13}$ and $D_{23}$ in Appendix~\ref{Append_Leg}. Thus, it was efficient to use the motion of $\theta_2$ as a virtual control input to balance $\theta_3$. When implementing the PEIC-based controller with $q_{au}=\theta_1$, the system cannot achieve the desired performance and becomes unstable. We also implemented the proposed controller with the physical model. The control errors are included and listed in Table~\ref{Tab_Leg_Error}. Compared with the learning-based controllers, the model-based control resulted in larger errors. Since the mechanical frictions and other unstructured effects were not considered, the physical model might not capture and reflect the accurate robot dynamics. The results confirmed the advantages of the proposed learning-based control approaches.

\section{Conclusion}
\label{Sec_Conclusion}

This paper presented a new learning-based modeling and control framework for underactuated balance robots. The proposed design was an extension and improvement of the EIC-based control with a GP-enabled robot dynamics. The proposed new robot controllers preserved the structural design of the EIC-based control and achieved both tracking and balance tasks. The PEIC-based control re-shaped the coupling between the actuated and unactuated coordinates. The robot dynamics was transferred into a fully actuated subsystem and one reduced-order underactuated balance system. The NEIC-based control compensated for uncontrolled motion in a subspace. We validated and demonstrated the new control design on two experimental platforms and confirmed that stability and balance were guaranteed. The comparison with the physical model-based EIC control and the MPC design confirmed superior performance in terms of the error bound. Extension of the GP-based learning control design for highly underactuated balance robots is one of the ongoing research directions.

\vspace{-3mm}
\section*{Acknowledgments}

The work was supported in part by the US National Science Foundation (NSF) under award CNS-1932370.

\vspace{4mm}

\appendix
\noindent{\bf Appendices}
\vspace{-6mm}
\section{Proofs}

\subsection{Proof of Lemma~\ref{Lem_Uncontrolled_Motion}}
\label{proof_free_motion}

The system dynamics $\mathcal S$ under the control $\bm u^\mathrm{ext}$ is
\begin{equation}
\label{Eq_S_under_uext}
  \ddot{\bm q}_a={\bm v}^\mathrm{ext},\;  \ddot{\bm q}_u= -\bm D_{uu}^{-1}(\bm D_{ua}{\bm v}^\mathrm{ext} +\bm H_{u}).
\end{equation}
When $\rank(\bm D_{au})=m$ holds for $\bm q$, the SVD in~\eqref{Eq_Dua_SVD} exists and all $m$ singular values are great than zero, i.e., $\sigma_i>0$. Thus $\ker(\bm D_{au})=\bm V_n$ contains $(n-m)$ column vectors. Plugging~\eqref{Eq_Dua_SVD} into~\eqref{Eq_S_under_uext} and considering the coordinate transformation, we obtain
\begin{equation}
 \ddot{\bm p}_a={\bm \nu}^\mathrm{ext},\; \ddot{\bm q}_u=-\bm D_{uu}^{-1}(\bm U\bm \Lambda_m \bm \nu ^\mathrm{ext}_{m} +\bm H_{u}), \label{Eq_Su_in_Vu}
\end{equation}
where $\bm U\bm \Lambda \bm V^T{\bm v}^\mathrm{ext}=\bm U\bm \Lambda_m \bm \nu ^\mathrm{ext}_{m}$ is obtained and used by using the fact that $\bm \Lambda \in \mathbb{R}^{m\times n}$ is a rectangular diagonal matrix.

Given the definition of $\mathcal{E}$, $\bm q_u^{e}$ is obtained by solving the algebraic equation $\bm \Gamma_0(\bm q_u;\bm v^\mathrm{ext})=\bm 0$. We replace $\bm D_{ua}(\bm q_u^{e})$ with $\bm D_{ua}(\bm q_u)$ in $\bm \Gamma_0$ and therefore, using~\eqref{Eq_Dua_SVD}, $\bm \Gamma_0=\bm 0$ is rewritten into
\begin{align}
\label{Eq_Gamma_BEM_Vu}
\bm \Lambda_m  \bm \nu_{m}^\mathrm{ext}+ \bm U^T \bm H_u^{gp}\Big\vert_{\bm q_u =\bm q_u^{e}, \dot{\bm q}_u= \ddot{\bm q}_u=\bm 0}=\bm 0.
\end{align}
The BEM $\mathcal{E}$ depends only on $\bm \nu_{m}^\mathrm{ext}$. That is, the control effect in $\ker(\bm D_{ua})$ is not used when obtaining the BEM.

Furthermore, since all controls show up in $\mathcal S_u$ dynamics, the control inputs should be updated and the EIC-based control in~\eqref{Eq_Ua_Int} exists. We substitute~\eqref{Eq_Dua_SVD} and~\eqref{Eq_Ua_Int} into $\mathcal S_a$ dynamics and obtain
\begin{align*}
\ddot{\bm q}_a=\bm v^\mathrm{int}=-\bm{D}_{u a}^{+}(\bm H_{u}+ \bm D_{uu} \bm v_u^{\mathrm{int}})=-\bm V \bm \Lambda^{+} \bm U^T(\bm H_{u}+ \bar{\bm D}_u \bm v_u^{\mathrm{int}}).
\end{align*}
Multiplying the above equation on both sides with $\bm V^T$ and considering~\eqref{Eq_transform}, $\mathcal S$ under the EIC-based control becomes~\eqref{Eq_Su_Vu2} and the $(n-m)$ coordinates are free of control.

\subsection{Proof of Lemma~\ref{Lem_PEIC}}
\label{proof_PEIC}

Under input $\bm u_u$, $\ddot{\bm q}_{au}= \bm v_{au}^\mathrm{int}$, we solve $\ddot{\bm q}_u$ by~\eqref{Eq_Su},
\begin{align*}
\ddot{\bm q}_u & =-\bar{\bm D}_{u u}^{-1}[\bar{\bm D}_{ua}^u \bm v_{au}^\mathrm{int}+\bm H_{un}]\nonumber\\
    &=-\bar{\bm D}_{u u}^{-1}[\bm H_{un} -\bar{\bm D}_{ua}^u(\bar{\bm{D}}_{u a}^u)^{-1}(\bm H_{un}+\bar{\bm D}_{uu} \hat{\bm v}_u^\mathrm{int})]=\hat{\bm v}_u^\mathrm{int}.
\end{align*}
Clearly, the unperturbed subsystem $\mathcal S^{gp}_{u}$ remains the same as that under the EIC-based control. With the designed control, the $\bm q_{aa}$ dynamics is unchanged and $\ddot{\bm q}_{aa} = \hat{\bm v}_a^\mathrm{ext}$ holds regardless of $\hat{\bm v}_u^\mathrm{int}$. For $\bm q_{aa}$ and $\bm q_{au}$, we obtain $\ddot{\bm q}_{aa} = \hat{\bm v}_{a}^\mathrm{ext}$ and $\ddot{\bm q}_{au}= \hat{\bm v}_u ^\mathrm{int}$. The relationship in~\eqref{Eq_Su_Vu2} indicates that if the unactuated subsystem dynamics is written into~$\ddot{\bm q}_u=\bm v_u^\mathrm{int}$, the dynamics $\ddot{\bm q}_a $ under the transformation $\bm \Upsilon$ must contain the portion~\eqref{Eq_Su_Vu2-a}. Similarly, we obtain
\begin{subequations}\label{Eq_S_PEIC}
\begin{align}
\hspace{-2.5mm}\mathcal S_\mathrm{PEIC}:~&\ddot p_{ai}=-\frac{\bm u_{i}^T\left(\bm H_u^{gp}+\bar{\bm D}_{uu}  \hat{\bm v}_u^\mathrm{int}\right)}{\sigma_{i}},~i=1,\cdots,m, \\
&\ddot p_{aj}=\bm v_{j}^T \hat{\bm v}_{a}^\mathrm{int},\; j=m+1,\cdots,n,\\
&\ddot {\bm q}_{u} = \hat{\bm v}_{u}^\mathrm{int},
\end{align}
\end{subequations}
where $\hat{\bm v}_{a}^\mathrm{int}=\left[(\hat{\bm v}_{a}^\mathrm{ext})^T~(\hat{\bm v}_u^\mathrm{int})^T\right]^T$. Since $\hat{\bm v}_{a}^\mathrm{int}$ is not obtained in the way as in~\eqref{Eq_Va_Int}, i.e., $\hat{\bm v}_{a}^\mathrm{int} \notin \ker(\bar{\bm D}_{ua})$,  $\bm v_{m+j}^T \hat{\bm v}_{a}^\mathrm{int} \neq \bm 0$ and $\bm p_{an}$ is under active control design.  Meanwhile $\bm v_{m+j}^T \hat{\bm v}_{a}^\mathrm{int}$ drives $\bm q_a\rightarrow \bm q_a^d$ in $\ker(\bar{\bm D}_{au})$, given that $\hat{\bm v}_{a}^\mathrm{ext}$ and $\hat{\bm v}_u^\mathrm{int}$ are designed to drive $\bm q_a\rightarrow \bm q_a^d$. Therefore, if the unperturbed system under EIC-based control is stable, it is also stable under the PEIC-based control. This completes the proof.

\subsection{Proof of Lemma~\ref{Lem_NEIC}}
\label{proof_NEIC}

Under the NEIC-based control input, the $\mathcal S^{gp}_{a}$ becomes
\begin{equation}\label{Eq_qa_improved}
\ddot{\bm q}_a =\tilde{\bm v}_{a}^{\mathrm{int}}=\tilde{\bm v}^\mathrm{int}+ \tilde{\bm v}_{a}^\mathrm{ext}=-\bar{\bm D}_{u a}^{+}(\bm H_u^{gp}+\bar{\bm D}_{uu} \hat{\bm v}_u^\mathrm{int})+\bm V_{n} \bm \nu_{n}.
\end{equation}
Plugging above equation into $\mathcal S^{gp}_{u}$, we obtain
\begin{align*}
\ddot{\bm q}_u=&-\bar{\bm D}_{u u}^{-1}(\bar{\bm D}_{u a} \ddot{\bm q}_a+\bm H_u^{gp})=-\bar{\bm D}_{u u}^{-1}[-\bar{\bm D}_{u a}\bar{\bm D}_{u a}^{+}(\bm H_u^{gp}+\nonumber\\
    &\bar{\bm D}_{u u} \hat{\bm v}_u^\mathrm{int})+\bar{\bm D}_{u a}\bm V_{n}  \bm \nu_ {n}+\bm H_u^{gp}]=\hat{\bm v}_u^\mathrm{int}-\bar{\bm D}_{u u}^{-1}\bar{\bm D}_{u a} \bm V_{n} \bm \nu_{n}.
\end{align*}
Using the SVD form of $\bar{\bm D}_{ua}$ in~\eqref{Eq_Dua_SVD} and $ \bm \Lambda \bm V^T \bm V_{n}\equiv\bm 0$, the above equation is further simplified as
\begin{equation}\label{Eq_Su_NEIC}
  \ddot{\bm q}_u=\hat{\bm v}_u^\mathrm{int}-\bar{\bm D}_{u u}^{-1}\bm U \bm \Lambda \bm V^T \bm V_{n}\bm \nu_{n}  =\hat{\bm v}_u^\mathrm{int}.
\end{equation}
Clearly,  $\mathcal S^{gp}_{u}$ dynamics is unchanged compared to~\eqref{Eq_Su_Vu2}.

We further apply the transformation $\bm \Upsilon$ to $\bm q_a$ and apply SVD to $\bar{\bm D}_{u a}^{+}$. The $\mathcal S_{u}$ dynamics~\eqref{Eq_qa_improved} and~\eqref{Eq_Su_NEIC}  become
\begin{subequations}\label{Eq_S_NEIC1}
\begin{align}
\hspace{-2.5mm}\mathcal S_\mathrm{NEIC}:~&\ddot p_{ai}=-\frac{\bm u_{i}^T\left(\bm H_u^{gp}+\bar{\bm D}_{uu}  \hat{\bm v}_u^\mathrm{int}\right)}{\sigma_{i}},\;i=1,\cdots,m,\\
&\ddot p_{aj}=\nu_{nj},\; j=m+1,\cdots,n,\\
&\ddot {\bm q}_{u} = \hat{\bm v}_{u}^\mathrm{int}.
\end{align}
\end{subequations}
Compared to~\eqref{Eq_Su_Vu2}, we add the control $\tilde{\bm v}_{a}^\mathrm{ext}$ to drive $\bm q_a\rightarrow\bm q_a^d$ in the subspace $\ker(\bar{\bm D}_{ua})$. Therefore, if the system~\eqref{Eq_Su_Vu2} is stable, \eqref{Eq_S_PEIC} is also stable, as the $\bm p_{am}$ and $\bm q_u$ dynamics are unchanged.

\subsection{Proof for Theorem~\ref{Thm_Error_Bound}}\label{proof_error_bound}

We present the stability proof for the PEIC- and NEIC-based controls using the Lyapunov method.

\emph{PEIC-Based Control}: Plugging~\eqref{Eq_error_dyna1} into the $V_1=V$ and considering~\eqref{Eq_Q}, we obtain $\dot{V}_1=\bm e^T(\bm A^T\bm P+ \bm P\bm A)\bm e+2 \bm e^T\bm P\bm O_1=-\bm e^T\bm Q\bm e+\bm e^T\bm Q_\Sigma\bm e+2 \bm e^T\bm P\bm O_1$. where $\bm Q_\Sigma=(\bm A -\bm A_0)^T \bm P +\bm P (\bm A -\bm A_0)$. The bounded variance leads to the bounded eigenvalue of matrix $\bm Q_\Sigma$. Given the fact that $\bm Q_\Sigma=\bm Q_\Sigma^T$, the eigenvalues of $\bm Q_\Sigma$ are real numbers.

We note that $\bm Q_\Sigma$ is bounded and $\bm P, \bm Q$ are constant. The perturbation term $\bm O_1$ is bounded as shown in~\eqref{Eq_O1}. Then $\dot V_1$ is rewritten as
\begin{align*}
 \dot{V}_1\le & -[\lambda_{\min}(\bm Q)-\lambda_{\max}(\bm Q_\Sigma)]\norm{\bm e}^2+ 2\lambda_{\max}(\bm P)\norm{\bm e}(d_1\\
  &+d_2\norm{\bm e}) +2\lambda_{\max}(\bm P)\norm{\bm e}\left(l_{u1}\norm{\bm \kappa_u}+l_{a1}\norm{\bm \kappa_a}\right)\\
 =& -[\lambda_{\min}(\bm Q)-\lambda_{\min}(\bm Q_\Sigma)-2d_2\lambda_{\max}(\bm P)]\norm{\bm e}^2\\
 &+ 2(d_1+\omega_1)\lambda_{\max}(\bm P)\norm{\bm e}.
\end{align*}
where $\omega_1=l_{u1}\norm{\bm \kappa_u}+l_{a1}\norm{\bm \kappa_a}$ denotes the uncertainties related to GP prediction errors. $\lambda_{\min}(\cdot)$ and $\lambda_{\max}(\cdot)$ denote the smallest and greatest eigenvalues of a matrix, respectively. Considering $\lambda_{\min}(\bm P)\norm{\bm e}^2 \le V_1\le \lambda_{\max}(\bm P)\norm{\bm e}^2$, we define
\begin{align*}
\gamma_1  =\frac{\lambda_{\min}(\bm Q)-\lambda_{\max}(\bm Q_\Sigma)-2d_2\lambda_{\max}(\bm P)}{\lambda_{\max}(\bm P)},
\end{align*}
$\rho_1 = 2d_1\lambda_{\max}(\bm P)\norm{\bm e}$, $\varpi_1  =2\omega_1\lambda_{\max}(\bm P)\norm{\bm e}$. With the bounded perturbations $\rho_1$ and $\omega_1$, the closed-loop system dynamics can be shown stable in the sense of probability as $\Pr\{V_1 \le -\gamma_1 V_1+\rho_1+\varpi_1\}>\eta$. Taking further analysis, we obtain a nominal estimation of the error convergence performance using the Lyapunov function as $\Pr\{\dot{V}_1 \le V_{1}(0)e^{-\gamma_1 t}\}>\eta$ and the error bound estimation $\Pr\{\|\bs{e}\| \le r_1\}>\eta$ with $r_1=\frac{2d_1\lambda_{\max}(\bm P)}{\lambda_{\min}(\bm Q)-\lambda_{\max}(\bm Q_\Sigma)-2d_2\lambda_{\max}(\bm P)}$.

\emph{NEIC-Basd Control}: Without the loss of generality, we select $\bm \nu_{n}=\bm V_{n}^T \hat{\bm v}^\mathrm{ext}$. We take $V_2=V$ as the Lyapunov function candidate for $\mathcal{S}_{e,\mathrm{NEIC}}$. If the control gains are the same as that in PEIC-based control and $\alpha=1$ for compensation effect,  $\gamma_2=\gamma_1$. We choose control gains properly such that $\gamma_2>0$. The system can be shown stable as $\Pr\{\dot{V}_2 \le -\gamma_2 V_2+\rho_2+\varpi_2\}>\eta$, where $\rho_2=2 d_1 \lambda_{\max}(\bm{P})\norm{\bm{e}}$, $\varpi_2=2 \omega_2 \lambda_{\max}(\bm{P})\norm{\bm{e}}$, and $\omega_2=l_{u2}\norm{\bm \kappa_u}+l_{a2}\norm{\bm \kappa_a}$ is defined same as $\omega_1$ containing the GP prediction uncertainties. A nominal estimation of error convergence and final error bound can also be obtained.

To show $\gamma_i>0$, $i=1,2$, the control gains should be properly selected. We assume that during the training data collection phase, the physical system is fully excited. The GP models are accurate compared to physical models. With a small predefined error limit as a stop criterion in BEM estimation, $c_i$ values can be shown as $c_i\ll 1$. Given the explicit form, $d_i$ are estimated for $\bm A_0$ and $\bm Q$, $\bm P$ is obtained by solving~\eqref{Eq_Q}. The matrix $\bm Q_\Sigma$ depends on the control gains associated with the prediction variance. Since the variance is bounded, we design $\bm k_{ni}$ such that $\lambda_{\max}(\bm Q_\Sigma)$ satisfies the inequality $\lambda_{\min}(\bm Q)-\lambda_{\max}(\bm Q_\Sigma)-2d_2\lambda_{\max}(\bm P)>0$. Thus the stability is obtained.

\section{Dynamics Model of Underactuated Balance Robots}

\subsection{Rotary inverted pendulum}
\label{Append_Rotary}
The dynamics model for the rotary pendulum is in the form of~(\ref{Eq_Physical_Model}) with $q_a=\theta_1$ and $q_u=\theta_2$. The model parameters are $\bm B=[1\;\, 0]^T$ and
\begin{align*}
    D_{aa}=&C(m_{p} l_{r}^{2}+0.25 m_{p} l_{p}^{2} \s^2_{2}+J_{r}),\\
    D_{au}=&D_{ua}=-0.5C{ m_{p} l_{p} l_{r} \c_{\alpha}}, \, D_{uu}=C{(J_{p}+0.25 m_{p} l_{p}^{2})},\\
    H_a=&C(0.5 m_{p} l_{p}^{2} \dot{\theta}_1 \dot{\theta}_2 \s_{2} \c_{2}+0.5 m_{p} l_{p} l_{r} \dot \theta_2^{2} \s_{\alpha}+\\
    & d_{r} \dot\theta_1+k_{g}^{2} k_{t} k_{m} \dot{\theta}_1/ R_{m}) +K_g k_t\dot\theta_2,\\
    H_u=&C(d_{p} \dot\theta_2-0.25 m_{p} l_{p}^{2} \c_{2} \s_2 \dot{\theta}^{2}-0.5 m_{p} l_{p} g \s_{2}),
\end{align*}
where $l_r$, $J_r$ and $d_r$ are the length, mass inertia and viscous damping coefficient of the base link, $l_p$, $J_p$ and $d_p$ are corresponding parameters of the pendulum, $m_p$ is the pendulum mass, $g$ is the gravitational constant, and $k_t, k_m, K_G, R_m, C$ are robot constant. The values of these parameters can be found in~\cite{Apk2011}. The control input is the motor voltage, i.e., $u=V_m$.

\subsection{3-link inverted pendulum}
\label{Append_Leg}
The model parameters for the 3-link inverted pendulum in~\eqref{Eq_Physical_Model} are
\begin{align*}
  D_{11} =& ( m_3( l_2^2 + 0.25l_3^2 ) + 0.25m_2l_2^2 - 0.5m_3l_3^2\c_3^2 - m_3l_2l_3\s_3 )\c _2^2  \\
   &+ ( 0.5m_3\s_3l_3^2 - m_3l_2l_3 )\s_2\c_3\c _2 + 0.25m_3\c_3^2l_3^2  \\
   &+ (0.25m_1 +m_2+m_3 )l_1^2 + J_1,\\
  D_{12} =& D_{21} =  - ( m_3l_2 + 0.5m_2l_2 )l_1\s_2 - 0.5m_3l_1l_3\c_{2 + 3},  \\
  D_{13} =& D_{31} =   0.5m_3l_1l_3\c_{2+ 3},  \\
  D_{22} =&J_2 + (m_3 + 0.25m_2)l_2^2 + 0.25m_3l_3^2 - m_3l_2l_3\s_3,  \\
  D_{23} =& D_{32} = ( 0.25l_3 - 0.5l_2\s_3)m_3l_3, \; D_{33}= J_3 + 0.25m_3l_3^2,\\
  G_1 = &0, \; G_2=  -(0.5m_2 + m_3 )\c_2l_2g + 0.5m_3l_3\s_{2+3}g,  \\
  G_3 = & -0.5m_3l_3\s_{2+3}g,
\end{align*}
where $m_i$, $l_i$ and $J_i$ are the mass, length, and mass inertia of each link, and $\s_{i+j}=\sin(\theta_i+\theta_j)$. Matrix $\bm C$ is obtained as $C_{ij}=\sum_{k=1}^3 c_{i j k} \dot{\theta}_k$, where Christoffel symbols $c_{i j k}=\frac{1}{2}\left(\frac{\partial D_{i j}}{\partial \theta_k}+\frac{\partial D_{i k}}{\partial \theta_j}-\frac{\partial D_{j k}}{\partial \theta_i}\right)$. The physical parameters are $m_1=0.7$~kg, $m_2=1.3$~kg, $m_3=0.3$~kg, $l_1=0.065$~m $l_2=0.23$~m, $l_3=0.25$~m, $J_1=0.0008$~kgm$^2$, $J_2=0.005$~kgm$^2$, $J_3=0.003$~kgm$^2$.

\bibliographystyle{asmems4}
\bibliography{HanRef}
\end{document}